\documentclass[a4paper,fleqn]{cas-dc}
\usepackage[authoryear]{natbib}
\usepackage[utf8]{inputenc}
\usepackage{times}
\usepackage{epsfig}
\usepackage{graphicx}
\usepackage{amsmath}
\usepackage{amssymb}
\usepackage{color}
\usepackage{multirow}
\usepackage{algorithmic}
\usepackage{algorithm}
\usepackage{placeins}
\usepackage{bbding} 
\usepackage{array}
\usepackage{textcomp}
\usepackage{stfloats}
\usepackage{url}
\usepackage{verbatim}
\usepackage{graphicx}
\usepackage{tablefootnote}
\usepackage{ulem} 
\usepackage{mathrsfs}
\usepackage[switch]{lineno}
\usepackage{subcaption}
\usepackage{anyfontsize}

\makeatletter
\renewcommand{\fnum@figure}{Fig. \thefigure.\@gobble}
\makeatother

\def\tsc#1{\csdef{#1}{\textsc{\lowercase{#1}}\xspace}}
\tsc{WGM}
\tsc{QE}
\makeatletter\def\Hy@Warning#1{}\makeatother
\begin{document}
\let\WriteBookmarks\relax
\def\floatpagepagefraction{1}
\def\textpagefraction{.001}

\shorttitle{Google Earth 3DGS Urban Reconstruction and Point Cloud Extraction}    

\shortauthors{G. et al.}

\title[mode = title]{Enhanced 3D Urban Scene Reconstruction and Point Cloud Densification using Gaussian Splatting and Google Earth Imagery}

\author[1]{Kyle-Yilin Gao}[orcid=0000-0002-8320-6308]
\ead{y56gao@uwaterloo.ca}
\credit{Conceptualization, Methodology, Validation, Formal analysis, Investigation, Data Curation, Writing - Original Draft, -Visualization, - Review \& Editing}

\affiliation[1]{organization={Department of Systems Design Engineering},
            addressline={University of Waterloo}, 
            city={Waterloo},
            state={Ontario},
            postcode={N2L3G1}, 
            country={Canada}}

\author[1]{Dening Lu}[orcid=0000-0003-0316-0299]
\credit{Investigation, Writing - Review \& Editing}
\ead{d62lu@uwaterloo.ca}

\author[2]{Hongjie He}[orcid=0000-0003-3839-5821]
\credit{Investigation, Writing - Review \& Editing}
\ead{hongjie.he@uwaterloo.ca}


\author[1]{Linlin Xu}[orcid=0000-0002-6833-6462]
\ead{l44xu@uwaterloo.ca}
\credit{Conceptualization, Resources, Writing - Review \& Editing, Supervision}
\cormark[1]
\cortext[1]{Corresponding authors. 
l44xu@uwaterloo.ca (Linlin Xu), junli@uwaterloo.ca (Jonathan Li)}

\author[1,2]{Jonathan Li}[orcid=0000-0001-7899-0049]
\ead{junli@uwaterloo.ca}
\credit{Conceptualization, Resources, Writing - Review \& Editing, Supervision}
\affiliation[2]{organization={Departmemt of Geography and Environmental Management},
            addressline={University of Waterloo}, 
            city={Waterloo},
            state={Ontario},
            postcode={N2L3G1}, 
            country={Canada}}
\cormark[1]

\begin{abstract}
Although large-scale remote sensing image based 3D urban scene reconstruction and modelling is crucial for many key applications such as digital twins and smart cities, it is a difficult task due to the uncertainties in heterogeneous datasets and the geometry models. This paper presents a Gaussian splatting based approach for 3D urban scene modeling and geometry retrieval, with the following contributions. First, we develop and implement a 3D Gaussian splatting (3DGS) approach for large-scale 3D urban scene modeling from heterogeneous remote sensing images. Second, we design point cloud densification approach in the proposed 3DGS model to improve the quality of 3D geometry extraction of urban scenes. Leveraging Google Earth imagery of different sensors, the proposed approach is tested on the region of University of Waterloo, demonstrating that the proposed approach greatly improves reconstructed point clouds quality over the other Multi-View-Stereo approaches. Third, we design and conduct extensive experiments on multi-source large-scale Google Earth remote sensing images across ten cities to compare the 3DGS approach with neural radiance field (NeRF) approaches, demonstrating improved view-synthesis results that greatly outperform previous state-of-the-art 3D view-synthesis approaches. 
\end{abstract}

\maketitle

\section{Introduction}
\label{sec:intro}
3D reconstruction and modelling from 2D images have recently received great interest given recent advances in photorealistic view synthesis methods with 3D reconstruction capabilities. From a technical perspective, it is an interdisciplinary research area spanning computer vision, computer graphics and photogrammetry. It finds applications in multiple domains, including autonomous navigation aided by 3D scene understanding \citep{2021self}, remote sensing and photogrammetry for crafting 3D maps essential for navigation, urban planning, and administration \citep{2015applications}. Moreover, it extends to geographic information systems incorporating urban digital twins \citep{2020twin,2020twin2}, as well as augmented and virtual reality platforms integrating photorealistic scene reconstructions \citep{2014ar,2022ar}.

This paper focuses on remote sensing-based large-scale view synthesis based on 3D Gaussian Splatting (3DGS), as well as 3D geometry extraction from 3D geometry extraction from Gaussian Splatting. Using only images from Google Earth Studio, we train a 3D Gaussian splatting model which outperforms previous NeRF-based models. We quantify and benchmark the view synthesis performance on a large-scale urban dataset with 10 cities captured from Google earth, as well as our region of study. We also extract the and densify the 3D geometry of the region of study using 3DGS, which we compare against a Multi-View-Stereo dense reconstruction. To our knowledge, this is the first use of 3D Gaussian Splatting for large-scale remote sensing-based 3D reconstruction and view-synthesis. 
\section{Back Ground and Related Work}
\label{sec:related}

\subsection{Urban 3D Photogrammetry}
Photogrammetry extracts 3D geometry and potentially other physical information from 2D images. Remote sensing-based urban photogrammetry for 3D city modelling relies on drones/aerial platforms/satellites whereby buildings of interest at captured at an oblique/off-nadir angle. This is often referred to as oblique photogrammetry. In large scale scenes, other land-uses and land-cover may be present and present additional challenges. Ground-based and airborne LiDAR scanners can also be used to generate very accurate 3D models, sometimes in conjunction with image-based methods. However, images are in general more accessible both in terms of sensor and data availability.

Traditional (non-deep learning-based) methods which generates 3D point cloud/geometry from images are grouped into two types: Structure-from-Motion (SfM) which generates sparse point clouds, and Multi-View-Stereo which generates dense point clouds \citep{2013urban3dsurvey}. The most fundamental method is perhaps Structure-from-Motion, which relies on multi-view geometry and projective geometry to establish the relationship between 3D points and their 2D projection onto imaging planes. Key points are extracted in each 2D image, and matched in images with scene overlap, then triangulated to three dimensions, and typically further calibrated/error-corrected using bundle adjustments or other methods, resulting in a sparse point cloud 3D reconstruction. The sparse point cloud can then be meshed and/or turned into digital surface models. Sparse SfM photogrammetry is typically applied as a preprocessing step, as shown in various works \citep{ 2015oblique_dense,2017obliqueUAV} to help with further dense reconstruction or data fusion with 3D scanned point clouds. Sparse SfM point clouds can only retrieve scene geometry, and cannot reproduce realistic 3D lighting of the scene which is crucial for AR/VR-based applications, and other applications which heavily rely visualization.

In urban settings, Multi-View-Stereo (MVS) also require oblique imagery to capture the geometry of buildings and their facades. Fundamentally, Multi-View-Stereo differs from sparse SfM photogrammetry since MVS aims for a dense reconstruction by making use of 3D information in each pixel of the 2D images, as opposed to specific key points in the 2D images. This can be done using various methods such as plane sweeping or stereo vision and depth map fusion, or even deep learning methods. MVS methods are typically divided into two categories volume-based, and point cloud based \citep{2013urban3dsurvey,mvssurvey}. Various authors \citep{2015oblique_dense,2017oblique,2017obliqueUAV,2020oblique,2022mvs,2024mvs} have employed MVS for dense urban 3D reconstruction which can also be meshed for various purposes such as digital surface modelling and geophysics simulations. However, compared to sparse SfM photogrammetry, dense MVS photogrammetry is much more computationally intensive, especially in terms of memory. Additionally dense MVS photogrammetry typically require sparse SfM photogrammetry, or at least the camera poses which is typically obtained from sparse SfM photogrammetry, as a preprocessing step. Dense reconstructions, although more visually appealing than sparse reconstructions, are still not photorealistic since they cannot model the directional dependence of lighting in the scene.

\subsection{Neural Radiance Fields and Urban 3D Reconstruction/View synthesis}
In recent years, Neural Radiance Field-based methods (NeRF) \citep{2021nerf} have dominated the field of novel view synthesis. Trained on posed images of a scene, NeRF methods use a differentiable rendering process to learn implicit \citep{2021mipnerf,2022mip360} or hybrid scene representation \citep{2022instantngp} typically as density and directional color fields, and typically using some Multi-Layer-Perceptron (MLP). The scene representation is then rendered into 2D images using a differentiable volume rendering process, allowing for scene representation learning via pixel-by-pixel supervised learning using back-propagation of a photometric loss. Certain explicit scene representation models \citep{2021plenoctrees,2022plenoxels,2022tensorf} use almost identical differentiable rendering pipelines, but store their scene representations explicitly, forgoing the use of decoding MLPs (although some of these methods allow for the use of a shallow decoding MLP, blurring the line between explicit and hybrid scene representation). 

To synthesize images, NeRF methods employ differentiable volume rendering, generating pixel color $C$ via alpha blending of local colors $c_i$ using local densities $\sigma_i$ along a ray with sampling intervals $\delta_i$. This is given by

\begin{equation} \label{eq:alpha_blending}
C = \sum_i c_i \alpha_i T_i
\end{equation}
where $c_i$ and $\sigma_i$ are sampled from the learned radiance field (e.g. the NeRF MLP), and 
\begin{equation}
    \alpha_i = 1 - \exp(-\sigma_i \delta_i) \; \text{and} \; T_i = \prod_{j=1}^{i-1} (1-\alpha_j).
\end{equation}

Urban scenes, unbound, full of transient objects (such as pedestrians, cars), and with changing lighting conditions pose a challenge to the learning of 3D scene representation. Methods such as NeRF-W \citep{2021nerfW}, Mip-NeRF360 \citep{2022mip360}, Block-NeRF \citep{2022blocknerf}, Urban Radiance Fields \citep{2022urbanrf} proposed solutions to some of these problems, and are suited for ground-level view-synthesis and 3D urban reconstruction.

Aerial view 3D reconstruction and view synthesis from remote sensing images was also attempted with methods such as Bungee/City-NeRF \citep{2022bungeenerf}, Mega-NeRF \citep{2022mega}, Shadow NeRF \citep{2021shadownerf}, Sat-NeRF \citep{2022satnerf}. These methods attempt to solve problems such as piecing together local NeRFs into large scale urban scene, multi-scale city view synthesis, and shadow-aware scene reconstruction for high-rises.  BungeeNeRF \citep{2022bungeenerf} is of interest as we extract a Google Earth dataset from our region of study using a similar method.

\subsection{3D Gaussian Splatting} 
\begin{figure*}[h]
\includegraphics[width=0.9\textwidth]{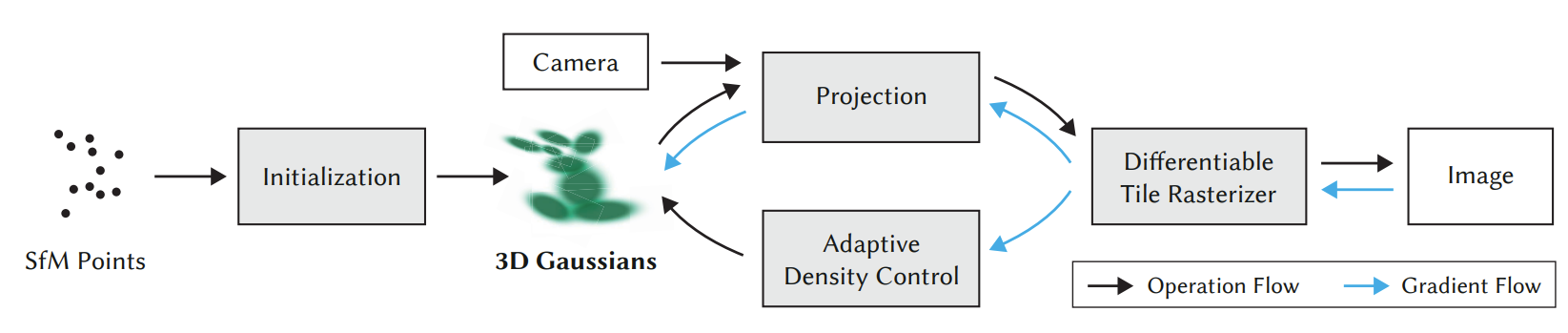}
\caption{3D Gaussian Splatting workflow, image from \citep{2023gaussian_splatting}. SfM is used to create a sparse point cloud to initialize the 3D Gaussian Splatting model. From these 3D Gaussians, new images are generated via the rasterizer and compared to ground truth images during optimization. Gaussians are densified as required.}\label{img:3dgsworkflow}
\end{figure*}

3D Gaussian Splatting (3DGS) \citep{2023gaussian_splatting} was first developed in 2023 as a view synthesis method competing against existing NeRF view synthesis methods. Compared against the vanilla NeRF method, the vanilla Gaussian Splatting method learns the 3D scene and synthesizes novel views orders of magnitude faster, and achieves a visual quality for view synthesis comparable and often exceeding to the best NeRF models, at the cost of a much larger memory footprint and requiring structure-from-motion (SfM) \citep{2016COLMAP} initialization/preprocessing. The workflow is visualized in Figure \ref{img:3dgsworkflow}

The SfM preprocessing is exactly the standard sparse photogrammetric process which identifies 2D key points, matches overlapping images, triangulates key points into 3D, and error-corrects trough bundle adjustment or some other method. Compared to standard photogrammetry which can sometimes project image colors into flat(lighting-less) 3D point cloud color, 3DGS is able to photorealistically reproduce the directionally dependent lighting of the scene, which is crucial for many applications. It is also able to fine-tune the geometry of the scene using photometric (color-based) objectives against ground truth pictures, as opposed to only minimizing re-projection errors in photogrammetry. Compared to NeRF models, 3DGS produces more natural 3D geometry, with natural correspondence between learned 3D positional means of 3D Gaussian functions and 3D point cloud representation of the scene geometry.


Representing the scene as 3D Gaussians functions, and representing lighting as spherical harmonic (SH) coefficients attached to these Gaussians, 3D Gaussian Splatting methods produces 2D images via a differentiable tile-based Gaussian rasterizer; projecting Gaussians into two dimensions based on novel view poses' view cones, alpha-blending the projected Gaussians to produce per-pixel color in the novel view. The novel views are supervised against ground truth images for training of Gaussian Splatting parameters. To the best of our knowledge, this is the first work to attempt large-scale remote sensing-based 3D reconstruction and view synthesis using 3D Gaussian Splatting, however recent works \citep{2024hierarchical,2024hugs} have applied Gaussian Splatting to large-scale urban street-level datasets.

\section{Method}
\label{sec:method}

\subsection{Region of Study}
The region of study is the Kitchener-Waterloo Region in Ontario, Canada, centered at the University of Waterloo. The city of Waterloo has a population of approximately 121 000 according to the 2021 census, and occupy 64.06 $\text{km}^2$ of land \citep{statcan}. The University of Waterloo lies at 43.472°N, 80.550°W, its main campus occupies 4.50 $\text{km}^2$. At the city scale, the study area is comprised of various land use and land cover features such as urban roads, buildings, agriculture, and other land use, low vegetation, water, mixed temperate forest, and other land cover. The study area is centered at the Environment-1 (EV-1) building located at 43.468°N, 80.542°W, and covers roughly an area of 165 km$^2$. We perform large-scale view synthesis at the city scale, and 3D point cloud comparison at the neighborhood scale. The Google Earth images retrieved for the scene are primarily from Landsat/Copernicus, Airbus, Data Scripps Institution of Oceanography (SIO), and National Oceanic and Atmospheric Administration (NOAA). 

The University of Waterloo lies on the traditional land of Neutral, Anishinaabeg and Haudenosaunee peoples. The University of Waterloo is situated on the Haldimand Tract, the land promised to the Six Nations that includes six miles on each side of the Grand River.

\subsection{Google Earth Studio Datasets}
\begin{figure*}[htbp]
\centering
    \centering
    \begin{subfigure}[b]{0.32\textwidth}
\includegraphics[width = 1\textwidth]{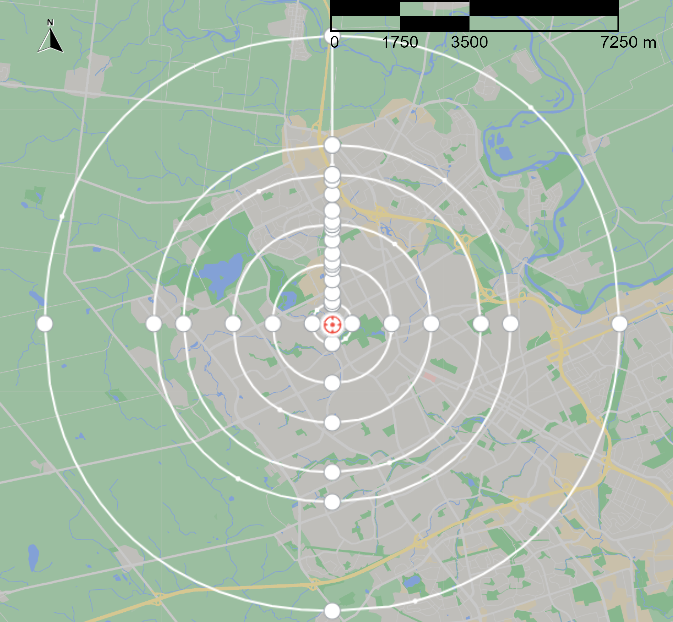}
    \end{subfigure}   
    \begin{subfigure}[b]{0.326\textwidth}
\includegraphics[width = 1\textwidth]{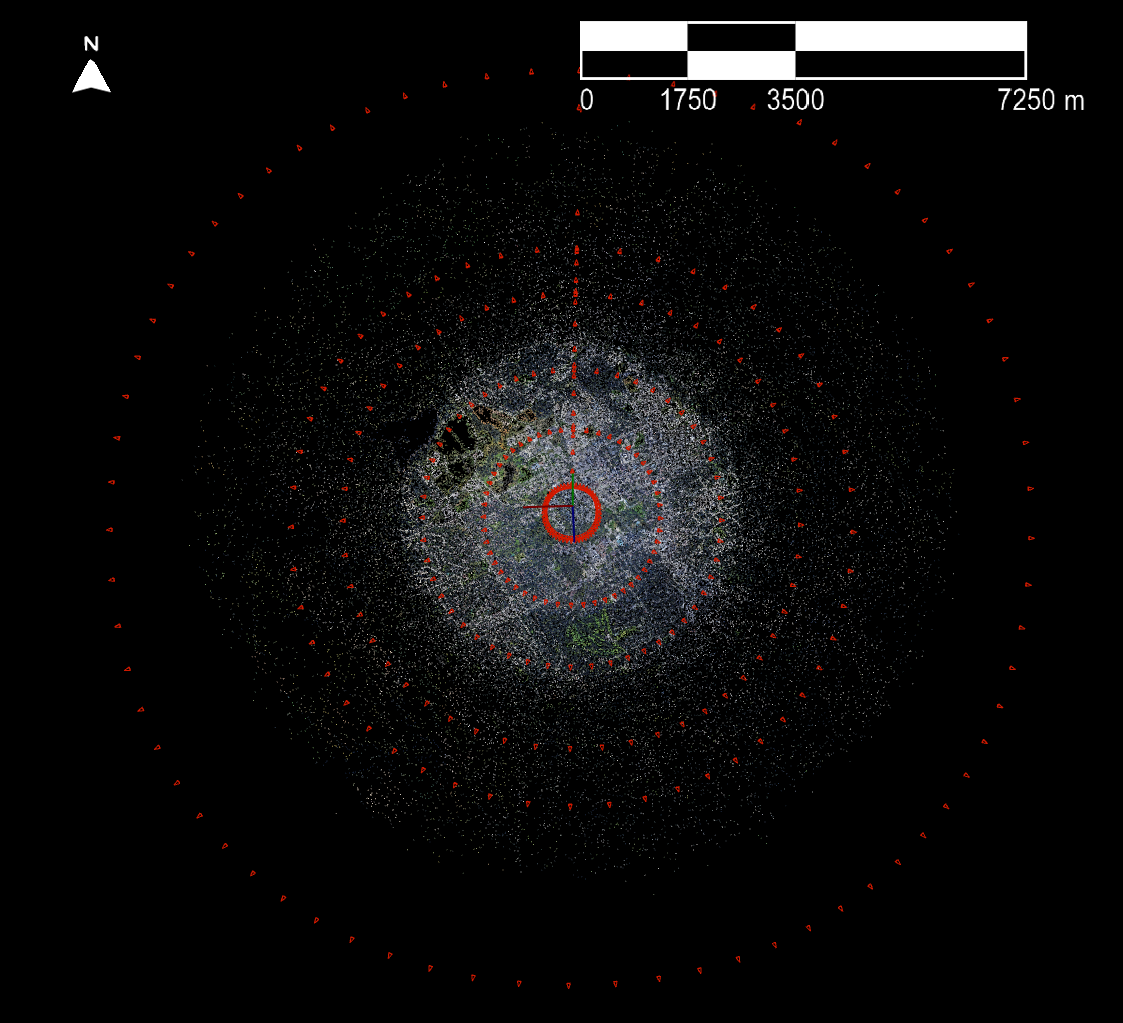}
    \end{subfigure}
        \begin{subfigure}[b]{0.32\textwidth}
\includegraphics[width = 1\textwidth]{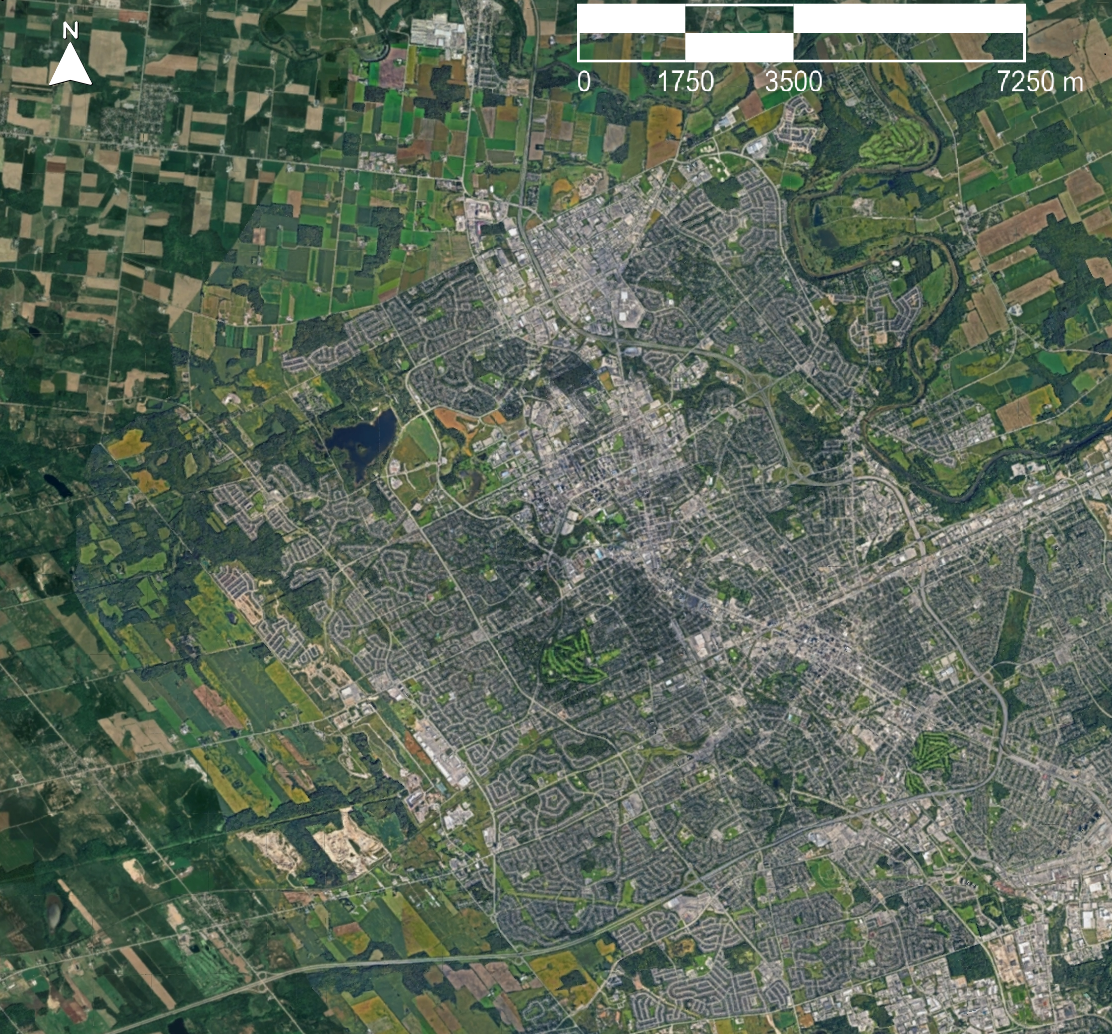}
    \end{subfigure}
\caption{Plot of camera poses, and reconstruction for region of study: Waterloo scene, centered on the EV1 building at the University of Waterloo.  \textbf{Left}: Google Earth Studio camera path. \textbf{Middle}: COLMAP \citep{2016COLMAP} SfM reconstructed sparse point cloud with projected color and camera poses (as red dots). \textbf{Right}: \href{https://www.google.com/help/terms_maps/}{Google Earth \copyright} \;aerial image of the region of study, sourced from Airbus. } \label{Fig:Uw_Scene}
\end{figure*}

For the region of study, as camera paths, we used seven concentric circles at different altitudes, radius and tilt angle centered around the EV-1 building at the University of Waterloo, Waterloo, Ontario, Canada. The first of these circles are of has a radius of 500 m, and an elevation of 475 m. The last circle has a radius of 7250 m and an elevation of 3690 m. All images point towards and above (at elevation 390 m) the EV-1 building in the University of Waterloo at 43.468 °N, 80.542 °W. The final circle's images have have a tilt angle of approximately 65.5 ° with respect to the horizontal with some deviations (within $\sim 0.3$ °). We gathered 401 images using Google Earth Studio along the camera path defined using these circles. The region of study and camera poses, along with sparse SfM results can be seen in Figure \ref{Fig:Uw_Scene}. During preprocessing, we observe poor SfM point cloud reconstruction results further than 6 km away from the scene center, reasonable SfM reconstruction within 6 km, and good SfM reconstruction within 1 km where individual buildings can be identified. The SfM preprocessing resulted in a sparse point cloud with 337382 points which were used to initiate 3D Gaussian functions for 3DGS. This multi-scale Google Earth Studio \citep{google_earth_studio} dataset was inspired by the BungeeNeRF dataset \citep{2022bungeenerf}, which we also use for a multi-city large-scale view-synthesis benchmark. 

For the BungeeNeRF scenes, we used the Google Earth Studio camera paths specified by BungeeNeRF \citep{2022bungeenerf}. The BungeeNeRF dataset consists of 10 scenes for 10 cities. Each scene is centered around a particular landmark, with camera paths defined by concentric circles of different orbit radius and elevation, with the scene coverage reaching city-wide at the highest elevation. Detailed information can found in Table \ref{tab:scenes} for the 10 BungeeNeRF scenes and the Waterloo scene. The New York scene centered at 56 Leonard and the San Francisco scene centered at Transamerica were used as main scenes for view reconstruction benchmark in BungeeNeRF \citep{2022bungeenerf}, and have 459 and 455 images respectively. These two scenes were rendered at 30 frames per second in a 1:30 minute video. All other scenes contain 221 images, and were rendered by fixing a frame limit of 220+1 given the fixed camera path, and were used for additional visualizations. We note the original BungeeNeRF paper contained two additional scenes (Sidney and Seattle), but the Google Earth Studio camera paths were not provided for these two scenes.

Google Earth Studio provides a platform for generating multi-view aerial/satellite images by simply specifying camera poses and scene location. Google Earth Studio produced composite images from various government and commercial sources, as well as rendered images from 3D models constructed using remote sensing images from these sources. These include Landsat/Copernicus, Airbus, NOAA, U.S. Navy, USGS, Maxar images and datasets taken at different times. An obvious example of composite image can be observed in the two bottom right images in Figure \ref{img:NY_SF}, with different Water color indicating different data sources and/or acquisition time.

\begin{table}[htbp]
\centering
\caption{Google Earth city scenes information} \label{tab:scenes}

\footnotesize
\begin{tabular}{lccc}
\toprule 
City & Landmark  & Lowest (m) & Highest (m) \\
\midrule

New York & 56 Leonard & 290 & 3,389 \\
San Francisco & Transamerica & 326 & 2,962 \\
Chicago & Pritzker Pavilion & 365 & 6,511 \\
Quebec & Château Frontenac & 166 & 3,390 \\
Amsterdam & New Church & 95 & 2,3509 \\
Barcelona & Sagrada Familia & 299 & 8,524 \\
Rome & Colosseum & 130 & 8,225 \\
Los Angeles & Hollywood & 660 & 12,642 \\
Bilbao & Guggenheim & 163 & 7,260 \\
Paris & Pompidou & 159 & 2,710 \\
\hline
Waterloo & EV-1 & 500 & 3690\\
\hline
\end{tabular}
\end{table}

\subsection{Structure from Motion Preprocessing and Sparse Point Cloud Extraction}
The standard implementation of 3D Gaussian Splatting relies on COLMAP \citep{2016COLMAP} for preprocessing. This SfM preprocessing takes a collection of unordered images with unknown camera poses, and outputs the camera pose of each image, as well as a sparse point cloud. Like all SfM methods, COLMAP SfM consists of the following steps.

\textbf{Feature Extraction}: In this step, for each image $I_i$, key points $x_j \in R^2$ are identified, and robust view invariant local features $f_j \in R^n$ are assigned. Scale Invariant Feature Transform (SIFT) features \citep{1999sift} are used as default in COLMAP, and provide robust features which allows for the same 3D point to be identified across multiple images as respective projected 2D key points.  
\textbf{Matching}: By searching through images and their respective features, potentially overlapping images pairs with matching key point features are identified.
    
\textbf{Geometric Verification}: A scene graph with images at nodes and edges connecting overlapping images is constructed by verifying potentially overlapping image pairs. This verification is done by estimating a valid homography in a potentially connected image pair using a robust estimation technique such as a variant of Random Sample Consensus (RANSAC) by \cite{1981ransac}. 

\textbf{Image Registration} From a starting image pair whose key points are triangulated into 3D, new images with overlap given the scene graph are added to the scene by solving the Pespective-n-Point problem \citep{1981ransac} which estimates camera poses given a number of 3D points and their 2D projection. This step robustly estimates the pose of the newly registered image. 

\textbf{Triangulation}: Given key points as viewed from two images with known poses, key points are triangulated \citep{2003multi-view} into 3D. Newly registered images extend the scene by allowing for more key points to be triangulated into the 3D reconstruction.

\textbf{Error-Correction}: To correct errors in registration and triangulation, bundle adjustment \citep{2000ba} is performed by jointly optimizing the camera poses $P_c \in SE3$ and 3D points $X_k \in R^3$ during the minimization of the reprojection loss $E$ given by the square error of the reprojection of the 3D point onto the image plane $\pi_{P_c}(X_k)$ with respect to pixel value $x_j \in R^2$. This is given by
    \begin{equation}
        E=\sum_j \rho_j (\pi_{P_c}(X_k)-x_j)^2.
    \end{equation}

 \cite{2016COLMAP} introduced various innovations improving the geometric verification, improving the robustness of the initialization and triangulation, introducing a next best view selection method and an iterative and more efficient bundle adjustment method, resulting in the COLMAP SfM library.

\subsection{Multi-View-Stereo Dense 3D reconstruction}
The MVS dense reconstruction we used as ground truth/reference geometry of the region of study is retrieved from COLMAP's MVS algorithm \citep{2016mvs}. This method is based on joint view selection and depth map estimation \citep{zheng}. The method is summarized as follows.

\textbf{Depth and normal map estimation}: To estimate the depth $\theta_l \in R^1$ and normal $n_l \in R^3$ at a pixel $l$ in a references image $X^{ref}$, a joint likelihood function is used. $\boldsymbol{X} = \{X^{ref},X^1,...X^m,...X^M\}$ is the collection of all images (with one image as source image, and the rest as reference images). $\boldsymbol{Z} = \{Z_l^m|l=1...L,m=1...M\}$ is the set of occlusion indicators, with $Z_l^m = 1$ if the image $X^m$ is selected for depth estimation of pixel $l$ in $X^{ref}$, and zero otherwise if occluded. $\boldsymbol{\theta} = \{\theta_l|l=1...L\}$ are the depths at each pixel $l$ of $X^{ref}$, which are to be recovered. $\boldsymbol{N} = \{n_l|l=1...L\}$ are the normals of $X^{ref}$ also to be recovered. This is given by 
\begin{align}
P(\boldsymbol{X,Z,\theta, N}) = \\ \prod_{l}\prod_{m}[P(Z^m_{l,t}|Z^m_{l-1,t},Z^m_{l,t-1})\\P(X^m_{l}|Z^m_{l},\theta_l,n_l)P(\theta_l,n_l|\theta^m_l,n^m_l)]
\end{align}
where $m$ indexes over input images, $l$ indexes pixels or patches in the references image $X^{ref}$, $t$ denotes optimization iteration. The first term $P(Z^m_{l,t}|Z^m_{l-1,t},Z^m_{l,t-1})$ enforces spatially smooth and temporally (in terms of optimization steps) consistent occlusion maps. The second term $P(X^m_{l}|Z^m_{l},\theta_l,n_l)$ enforces photometric consistency between the reference image and source images. The third term $P(\theta_l,n_l|\theta^m_l,n^m_l)$ enforces depth and normal maps consistent with multi-view geometry. Reader are referred to \cite{2016mvs} for the construction of each term this joint likelihood function and its optimization process.

\textbf{Filtering and fusion}: 
First, depth and normal maps for each image are estimated in accordance with the previous step. Photometric and geometric constraints are derived and used to filter outliers, where any observation $x^l$ who's support set $S_l = \{x^m_l\}$ satisfying less both geometric and photometric constraints has less than 3 elements (i.e. the reference pixel can be observed while satisfying both constraints in at least 3 other images.) A directed graph of consistent pixels is defined with supported pixels as nodes, edges pointing from reference to source image. The fusion is initialized at the node with maximum support (observed by the most source images while satisfying photometric and geometric constraints). Recursively, connected nodes are collected under a depth consistency constraint, a normal consistency constraint, and a reprojection error bound constraint. The collection's elements are fused when there are no more nodes satisfying all 3 constraints. The fused point becomes part of the output dense point cloud with location $p_j$ and normal averaged $n_j$ over the collection's elements. The fused nodes are culled from the graph and the process is repeated until the graph is empty. The final output is a dense point cloud with normals, which can be meshed via Poisson Surface Reconstruction \citep{2013poisson} as we have done, or using other methods if desired.

\subsection{3D Gaussian Splatting}
3D Gaussian Splatting \citep{2023gaussian_splatting}, which we briefly describe in this subsection, is used as foundation for our 3D urban reconstruction and view-synthesis experiments in the region of study and in the benchmarks. 

From 2D images of a scene, 3D Gaussian Splatting learns and represents the scene geometry as (unormalized) 3D Gaussian functions with mean $\mu \in R^3$ and $3 \times 3$ covariance matrix $\Sigma$ given by 
\begin{equation} \label{eq:Gaussian}
    G(x) = e^{-\frac{1}{2} (x-\mu)^T \Sigma^{-1} (x-\mu)}.
\end{equation}
The scene lighting and color are learned as third order spherical harmonics coefficients for each color channel attached to each Gaussian. To each Gaussian is also assigned a local (conic) opacity $\sigma$. Combined with the 3D mean and covariance matrix, resulting in a total of 59 trainable parameters per Gaussian. The 3D covariance matrix $\Sigma$ is learned as a 3D diagonal scaling matrix $S$, and a rotation represented by a quaternion $(r,i,j,k)$, which can be then used to reconstruct a 3D rotation matrix $R$ as follows

\begin{equation}
R = \begin{bmatrix}
    1 - 2(j^2 + k^2) & 2(i j - k  r) & 2(i  k + j  r) \\
2(i  j + k  r) & 1 - 2(i^2 + k^2) & 2(j  k - i  r) \\
2(i  k - j  r) & 2(j  k + i  r) & 1 - 2(i^2 + j^2)
\end{bmatrix}.
\end{equation}
The 3D covariance matrix is then given by
\begin{equation}
    \Sigma = RSS^TR^T.
\end{equation}

A sparse initial point cloud and training image camera poses are first computed using a structure from motion library such as COLMAP \citep{2016COLMAP}. A Gaussian is initialized at each point in the sparse point cloud, and trained using the differentiable tile-based rasterizer.

\subsubsection{Rasterization}

The tile-based rasterizer tiles the image into $16 \times 16$. For each tile, a view frustrum is projected into the 3D scene. 3D Gaussians are accumulated/assigned per-tile according to overlap to the view frustrum, and are projected into 2D and via projection of its covariance matrix $\Sigma$. Starting in homogenous coordinates, this is given by 
\begin{align}
    \Sigma' = J W \Sigma W^T J^T
\end{align} \label{eq:proj}where $W$ is the view transformation and $J$ is affine approximation of the projective transformation. The projective transformation is a matrix multiplication in homogenous coordinates in the case of a linear camera model such as the pinhole model used with the standard 3DGS model. In which case, $J$ is simply obtained from the intrinsic camera matrix. The third column of $\mu'$, and the third row and third column of $\Sigma'$ are then skipped to obtain the 2D mean and the 2D covariance matrix in the imaging plane in Cartesian coordinates.

 Gaussians are then sorted according to tile and by depth. For each pixel in a tile, the pixel's color is generated via alpha blending accumulating in-scene direction dependent color using the learned SH coefficients. For each Gaussian to be blended together, each $\alpha_i$ at a pixel location $x$ is given by evaluating the associated 2D Gaussian scaled by its associated learned opacity $a_i$.
 \begin{equation}
   \alpha_i(x) = a_i G_{2D}(x)
 \end{equation}
 where $G_{2D}(x)$ is the Gaussian (\ref{eq:Gaussian}) projected into 2D dimension and onto the image plane via (\ref{eq:proj}). 

 The rasterizer generates an image which is compared to the ground truth images using a photometric $L_1$ loss and $L_{D-SSIM}$, a difference of Structural Similarity Index Measure (D-SSIM) \citep{2004ssim}  loss via 
\begin{equation}
    L = (1 - \lambda)L_1 + \lambda_{D-SSIM}
\end{equation}
with $\lambda$ being an adjustable weighing parameter defaulting to 0.2. The trainable parameters are back-propagated through the differentiable rasterization and optimized using Adam \citep{2014adam}.

\subsubsection{Densification and Pruning}
\begin{figure}[h!]
\includegraphics[width=0.5\textwidth]{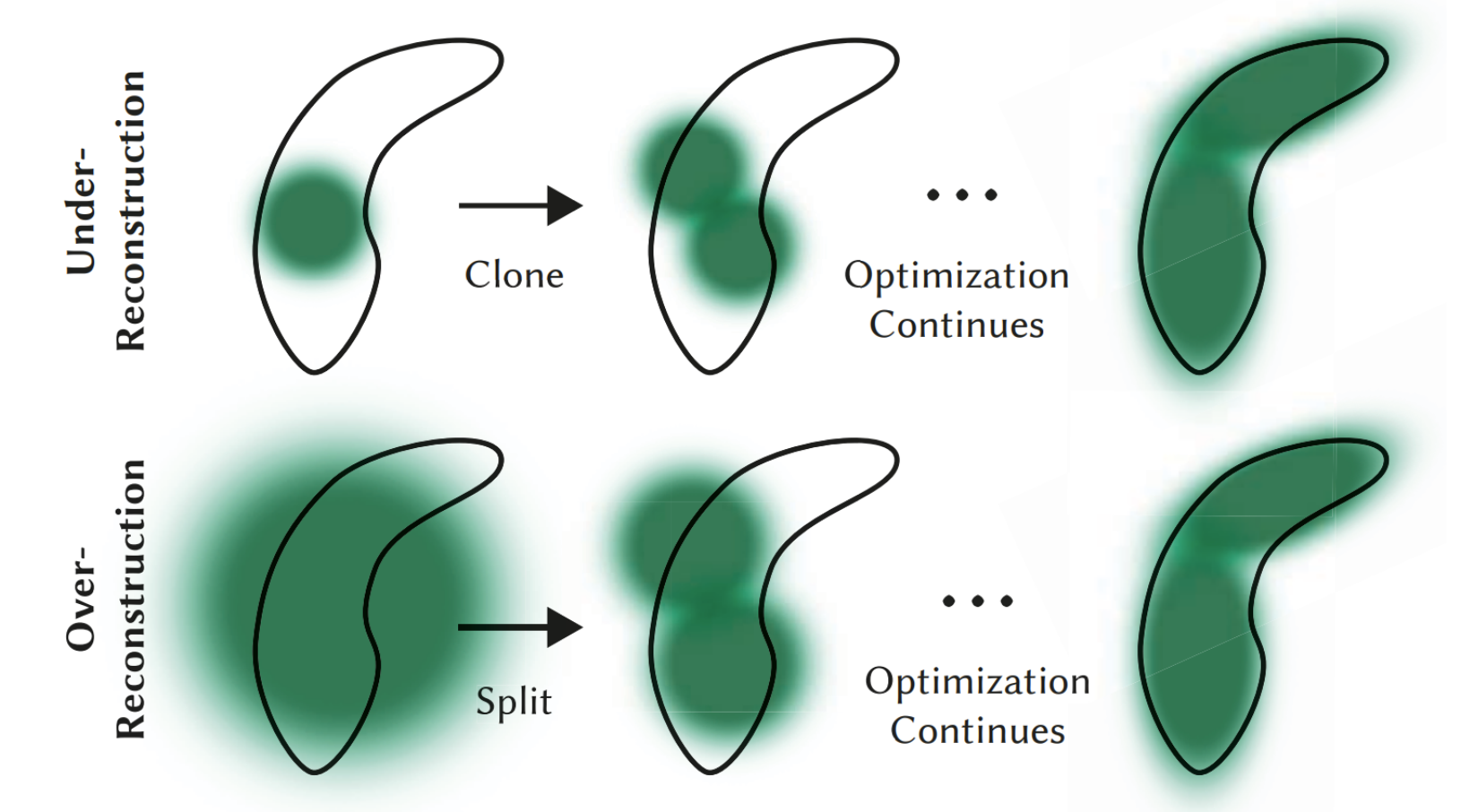}
\caption{Adaptive Gaussian Densification, adapted from \citep{2023gaussian_splatting}. \textbf{Top}: Cloning Gaussians to cover regions with under-reconstructed details. \textbf{Bottom}: Splitting Gaussians that cover large area but insufficiently represent local geometry.}\label{img:densification}
\end{figure}

3D Gaussian Splatting also densifies/grows new Gaussians in regions with high view-space positional gradient (threshold $\tau{pos}> 2.0 \times 10^{-4}$ as default). These regions correspond to neighborhoods with missing geometric features and regions with a few Gaussians covering large areas of the scene. Low variance Gaussians with view-space position gradients are duplicated. On the other hand, high variance Gaussians are split into two with standard deviation divided by a factor of 1.6. This is illustrated in Figure \ref{img:densification}.

Unimportant Gaussians are also pruned. Gaussian that are essentially transparent with opacity less than some user defined threshold ($a<\epsilon_a$, default value at $5e-3$) are deleted. Every 3000 iterations (or some other number of the user's choosing), every Gaussian's opacity is set to zero, then allowed to be re-optimized, then culled where needed. This process controls the number of floater artifacts, and helps control the total number of Gaussians. We believe that this densification and density control process can allow for point cloud reconstruction of similar density and potentially quality compared to a dense reconstruction, given a good dataset.  

\subsection{Evaluation Metrics}
For quality of synthesized images, we use Peak Signal to Noise Ratio (PSNR), Structural Similarity Index Measure (SSIM) \citep{2004ssim}, and Learned Perceptual Image Patch Similarity (LPIPS) \citep{2018lpips} as full reference image assessment metrics comparing generated views to ground truth views. PSNR is a good indicator of presence of noise and visual artifacts, whereas SSIM and LPIPS have been shown to better correlate with human judgement of the visual similarity of an image to its reference image. 




For point cloud geometry assessment, we used point-to-point (D1) mean squared error (MSE), point-to-surface (D2) MSE, Hausdorff distance, Chamfer distance, all of which compare a lower quality point cloud to its reference point cloud. We note that metrics such as D1 and D2 MSE do not penalize differences in point density, only deviations of existing points from ground truth/reference points. On the other hand, Chamfer and Hausdorff distances better capture the difference between the distribution of points, including differences in point density. 
\section{Experiments and Results} \label{sec:experiments}
\subsection{Experiment Setup}
Both COLMAP preprocessing and 3D Gaussian Splatting optimization were performed on a 3080 RTX GPU with 10GB VRAM, i9-10900KF CPU, with PyTorch version 2.1.1 and CUDA toolkit version 12.1. We note the GPU VRAM limitation being especially relevant, as it is always possible to grow more and more Gaussians to achieve higher and higher visual reconstruction quality at the cost of memory and storage while using 3D Gaussian Splatting.
\begin{figure*}[h!]
    \centering
    \begin{subfigure}[b]{0.32\textwidth}
    \includegraphics[width=\textwidth]{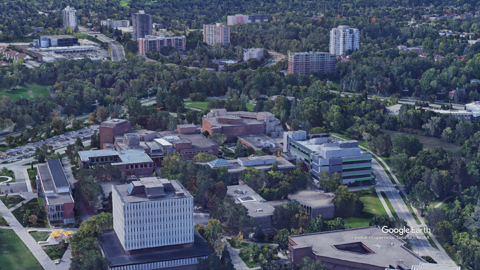}
    \end{subfigure}   
    \begin{subfigure}[b]{0.32\textwidth}
    \includegraphics[width=\textwidth]{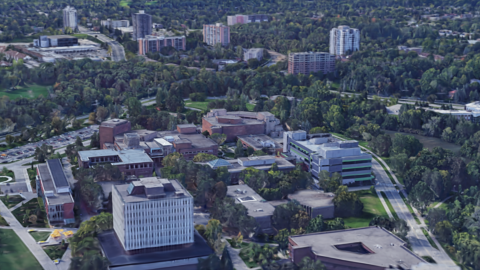}
    \end{subfigure}~
    \begin{subfigure}[b]{0.32\textwidth}
    \includegraphics[width=\textwidth]{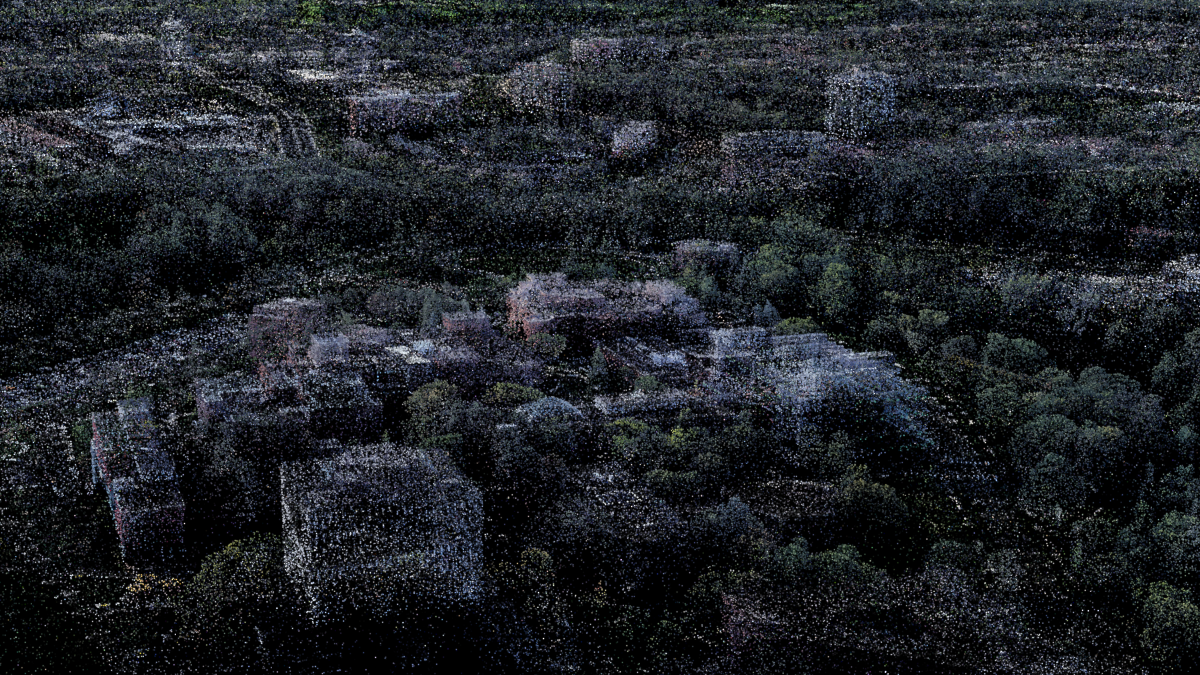}
    \end{subfigure}    
    
    \centering
    \begin{subfigure}[b]{0.32\textwidth}   \includegraphics[width=\textwidth]{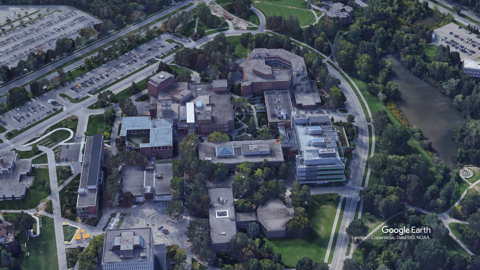}
    \end{subfigure}   
    \begin{subfigure}[b]{0.32\textwidth}    \includegraphics[width=\textwidth]{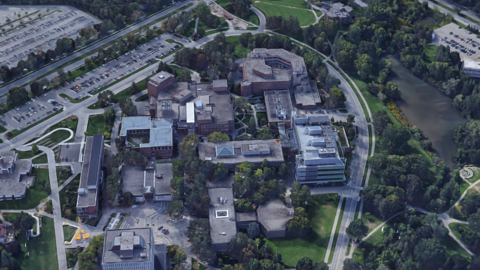}
    \end{subfigure}~
    \begin{subfigure}[b]{0.32\textwidth}    \includegraphics[width=\textwidth]{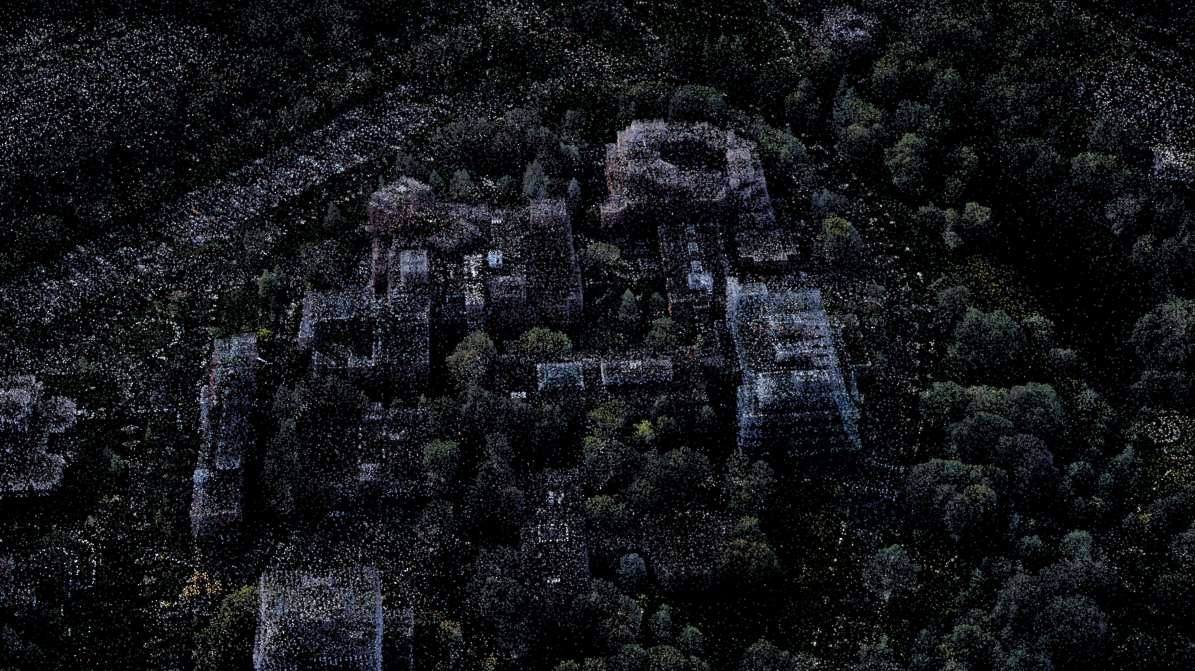}
    \end{subfigure}

    \centering
    \begin{subfigure}[b]{0.32\textwidth}   \includegraphics[width=\textwidth]{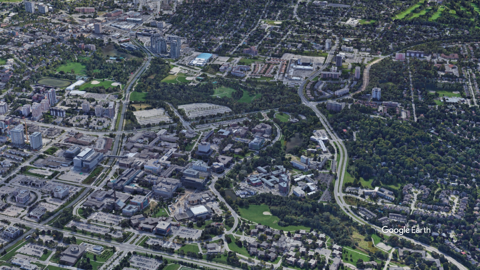}
    \end{subfigure}   
    \begin{subfigure}[b]{0.32\textwidth}    \includegraphics[width=\textwidth]{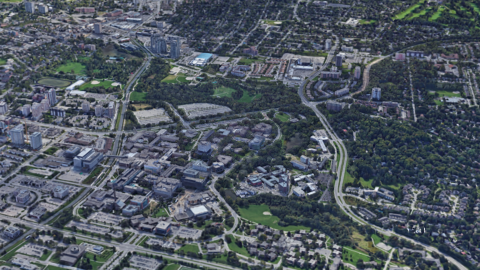}
    \end{subfigure}~
    \begin{subfigure}[b]{0.32\textwidth}    \includegraphics[width=\textwidth]{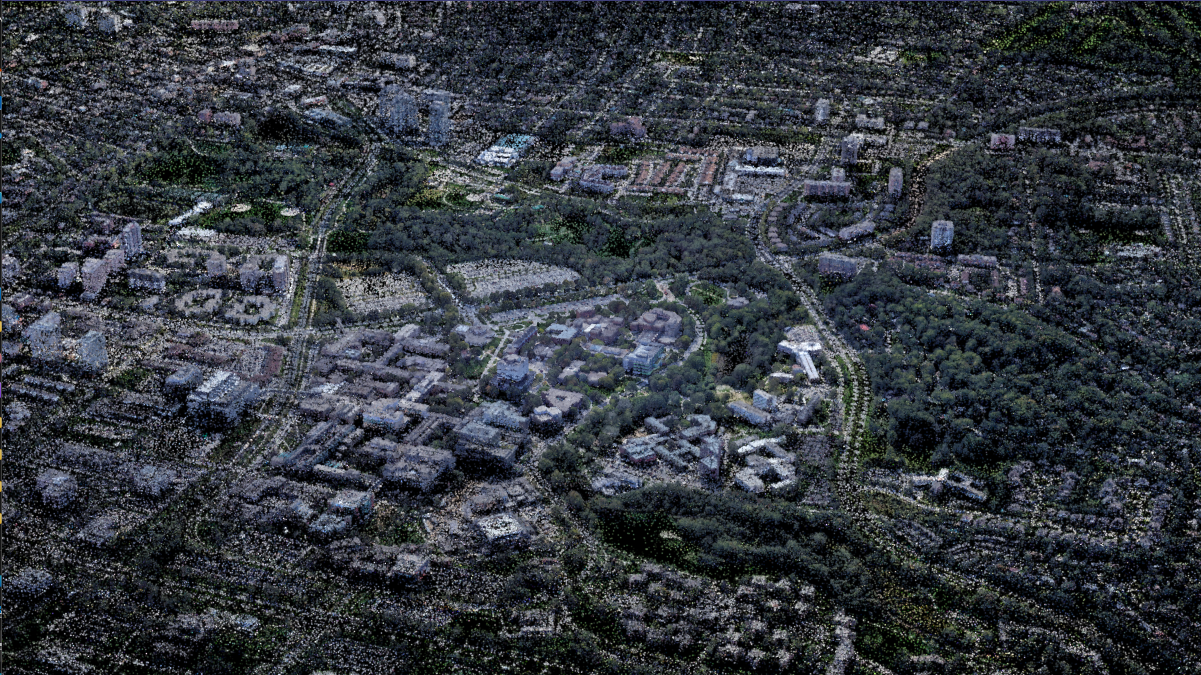}
    \end{subfigure}    

    \centering
    \begin{subfigure}[b]{0.32\textwidth}   \includegraphics[width=\textwidth]{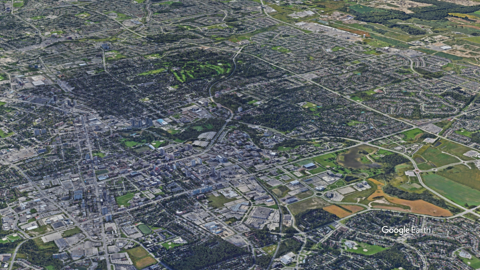}
    \end{subfigure}   
    \begin{subfigure}[b]{0.32\textwidth}    \includegraphics[width=\textwidth]{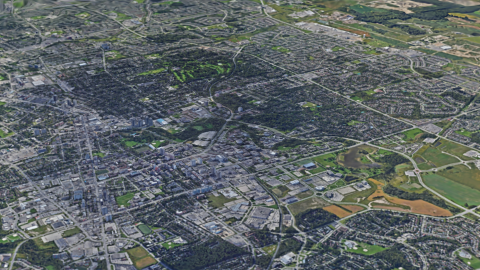}
    \end{subfigure}~
    \begin{subfigure}[b]{0.32\textwidth}    \includegraphics[width=\textwidth]{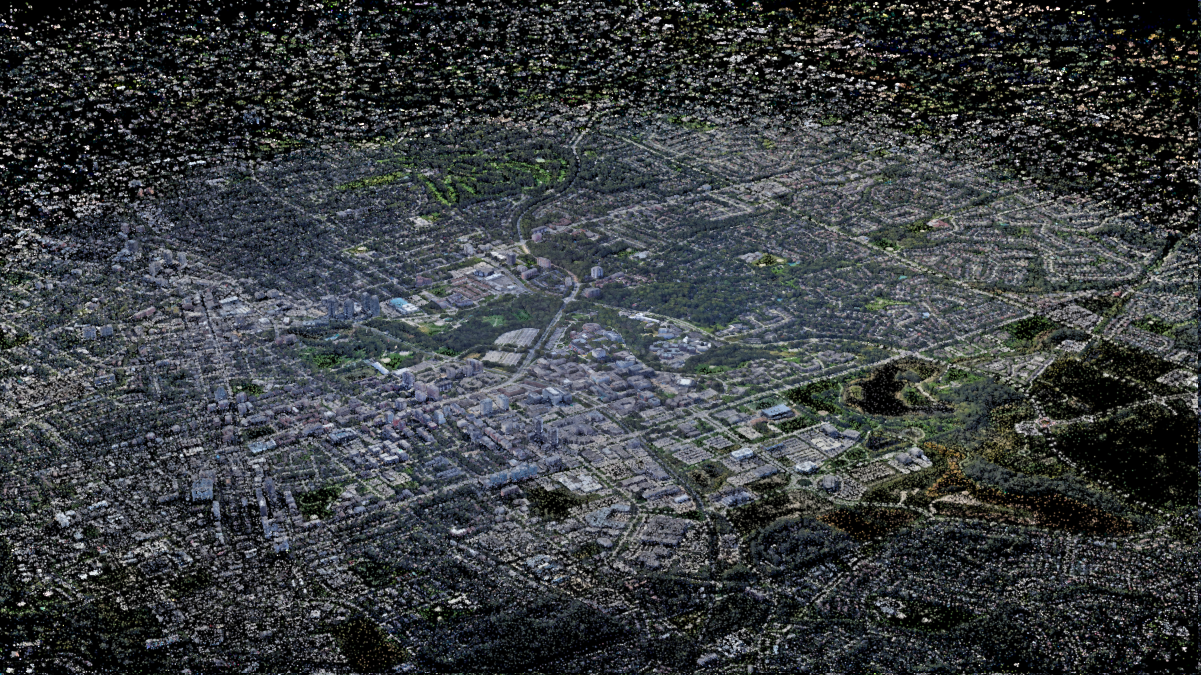}
    \end{subfigure}    

\caption{Ground truth, generated images, and visualization of Gaussian means of our Waterloo scene at different altitudes and orientations. \textbf{Left}: Waterloo scene ground truth; \textbf{Middle}: Waterloo scene 3DGS generated image; \textbf{Right} Waterloo scene visualization of location of each 3DGS Gaussian, i.e. 3D positional mean of each Gaussian. These points were then extracted as point clouds.} \label{img:uw}
\end{figure*}
\subsection{3D Novel View Synthesis of the Region of Study}

For the region of study, we used COLMAP SfM \citep{2016COLMAP} preprocessing and extracted 3D points and camera poses from the 400 two dimensional images. The experiments were performed with MipNeRF-360 \citep{2022mip360} style training validation split: one in eight images ($\sim$ 12.5\%) were reserved for testing purposes. The 1920 by 1080 resolution images were downsampled by a factor of 4 during training due to GPU memory constraints. We started densification at the 1000th iteration, and trained for 50000 iterations, densifying every 100 iterations. We used an initial positional learning rate of $3.2 \times 10^{-5}$ and a scale learning rate of $2 \times 10^{-3}$. The other training hyper parameters were kept as default. 

The results are shown in Table \ref{tab:psnr} and Figure \ref{img:uw}, in conjunction with further view synthesis experiments from the BungeeNeRF dataset. We achieve high view synthesis visual quality on both the training and test set. From visual inspection, the rendered images are nearly indistinguishable from the ground truth images. This is also supported by the visual assessment metrics, with SSIM scores near 1, LPIPS scores near 0, which indicates almost perfect visual agreement between ground truth and generated images. The PSNR values of around 30dB also are indicative of good image quality and low level of noise. This is comparable to the PSNR of a compressed image with respect to a full-sized image with a good lossy compression algorithm \citep{2013psnr_comp}, which is impressive considering the 3DGS model was trained at $\frac{1}{4}$ resolution.

\subsection{3D Novel View Synthesis of Bungee-NeRF Scenes}

 For the BungeeNeRF scenes, the experiments were performed with MipNeRF-360 style training and validation split as previously described. The experimental settings were kept the same as the Waterloo scene experiments, except we reduced the total number of training iteration to 30000. BungeeNeRF provided detailed benchmarks for the New York and San Francisco scenes, which were used for their main view-synthesis experiments. We performed detailed comparisons of Gaussian Splatting against BungeeNeRF, vanilla NeRF and Mip-NeRF for these two scenes. Additionally, we trained and evaluated Gaussian Splatting models for the remaining eight scenes whose camera paths were provided by BungeeNeRF.  

As observed in Table \ref{tab:2sceneBenchmark}, across both the New York and San Francisco scenes, we see a large increase in view synthesis quality according to all three metrics. The visual quality improvement from 3DGS to BungeeNeRF is much larger than the quality improvement from BungeeNeRF to vanilla NeRF or any other benchmarked models. We also note the large increase in view synthesis quality does not come at the cost of training time. In fact, the training of Gaussian Splatting models is three to four orders of magnitudes faster than implicit NeRF models such as vanilla NeRF \citep{2021nerf}, Mip-NeRF \citep{2021mipnerf}, and BungeeNeRF \citep{2021mipnerf}. Gaussian Splatting models achieve higher view synthesis quality with faster training and rendering time at the cost of memory and storage requirement \citep{2022nerfsurvey}. 

\begin{figure*}[htbp]
    \centering
    \begin{subfigure}[b]{0.24\textwidth}
    \includegraphics[width=\textwidth]{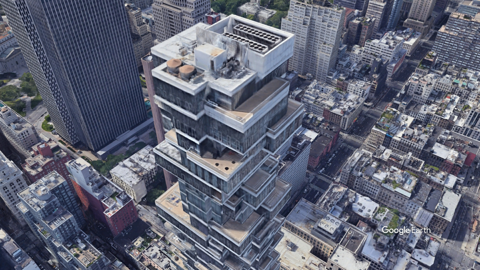}
    \end{subfigure}   
    \begin{subfigure}[b]{0.24\textwidth}
    \includegraphics[width=\textwidth]{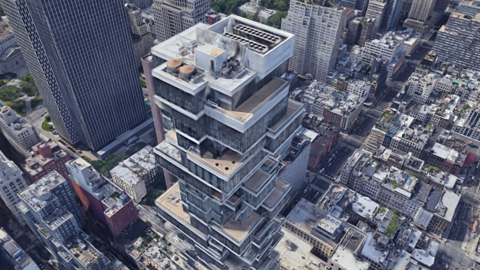}
    \end{subfigure}~
    \begin{subfigure}[b]{0.24\textwidth}
    \includegraphics[width=\textwidth]{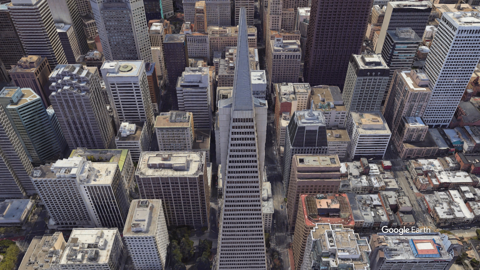}
    \end{subfigure}    
    \begin{subfigure}[b]{0.24\textwidth}
    \includegraphics[width=\textwidth]{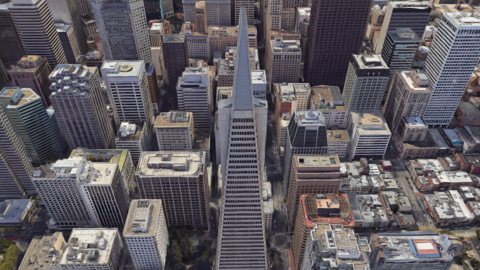}
    \end{subfigure}
    
    \centering
    \begin{subfigure}[b]{0.24\textwidth}   \includegraphics[width=\textwidth]{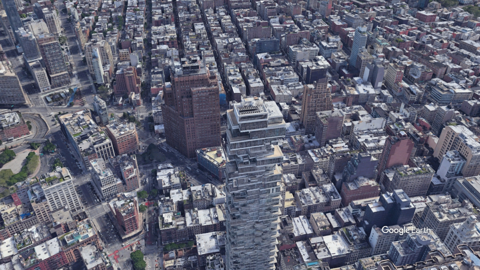}
    \end{subfigure}   
    \begin{subfigure}[b]{0.24\textwidth}    \includegraphics[width=\textwidth]{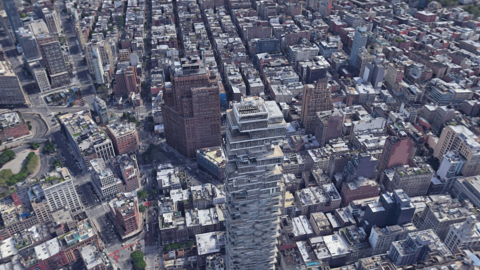}
    \end{subfigure}~
    \begin{subfigure}[b]{0.24\textwidth}    \includegraphics[width=\textwidth]{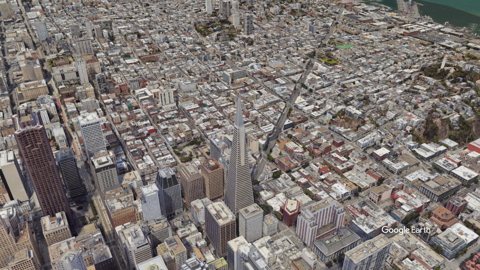}
    \end{subfigure}    
    \begin{subfigure}[b]{0.24\textwidth}    \includegraphics[width=\textwidth]{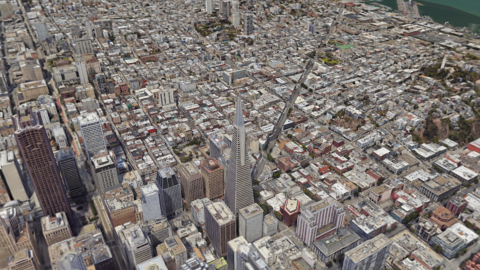}
    \end{subfigure}

    \centering
    \begin{subfigure}[b]{0.24\textwidth}   \includegraphics[width=\textwidth]{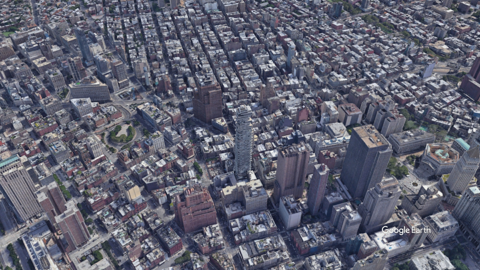}
    \end{subfigure}   
    \begin{subfigure}[b]{0.24\textwidth}    \includegraphics[width=\textwidth]{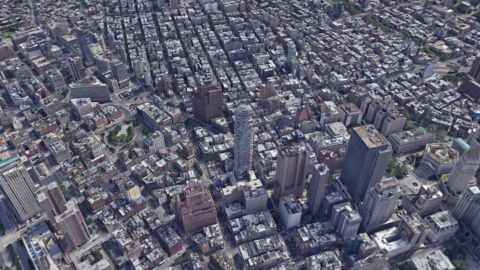}
    \end{subfigure}~
    \begin{subfigure}[b]{0.24\textwidth}    \includegraphics[width=\textwidth]{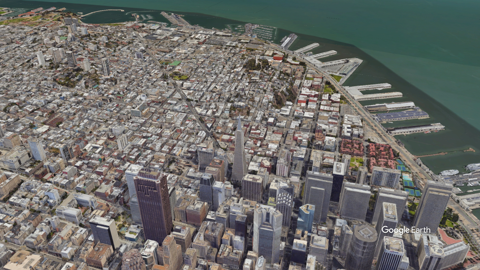}
    \end{subfigure}    
    \begin{subfigure}[b]{0.24\textwidth}    \includegraphics[width=\textwidth]{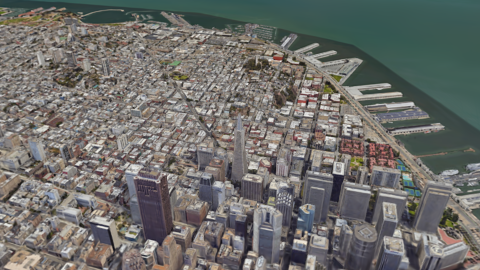}
    \end{subfigure}

    \centering
    \begin{subfigure}[b]{0.24\textwidth}   \includegraphics[width=\textwidth]{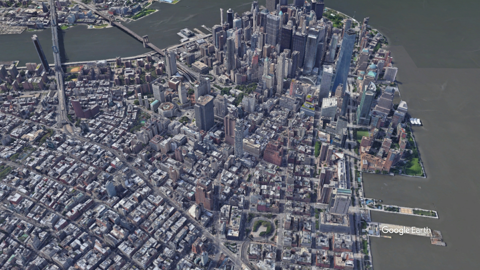}
    \end{subfigure}   
    \begin{subfigure}[b]{0.24\textwidth}    \includegraphics[width=\textwidth]{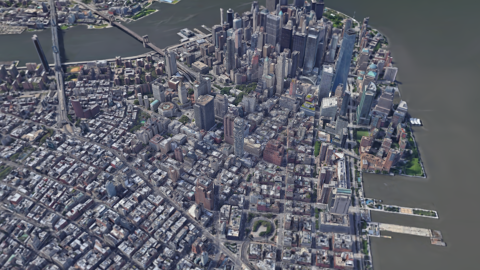}
    \end{subfigure}~
    \begin{subfigure}[b]{0.24\textwidth}    \includegraphics[width=\textwidth]{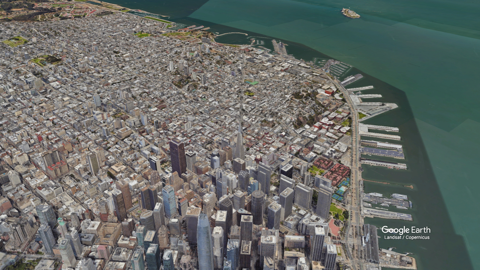}
    \end{subfigure}    
    \begin{subfigure}[b]{0.24\textwidth}    \includegraphics[width=\textwidth]{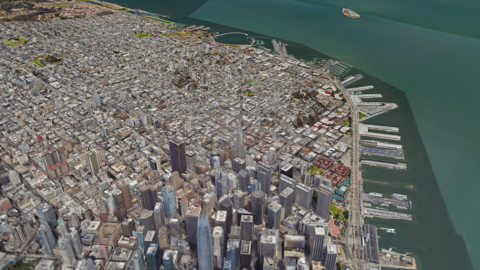}
    \end{subfigure}

    \centering
    \begin{subfigure}[b]{0.24\textwidth}   \includegraphics[width=\textwidth]{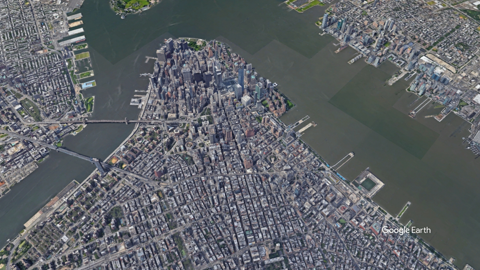}
    \end{subfigure}   
    \begin{subfigure}[b]{0.24\textwidth}    \includegraphics[width=\textwidth]{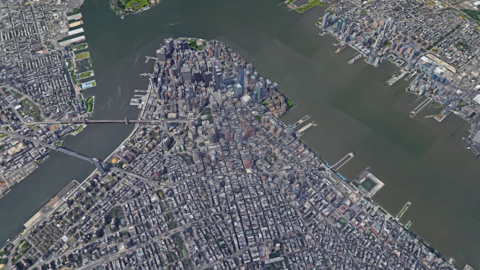}
    \end{subfigure}~
    \begin{subfigure}[b]{0.24\textwidth}    \includegraphics[width=\textwidth]{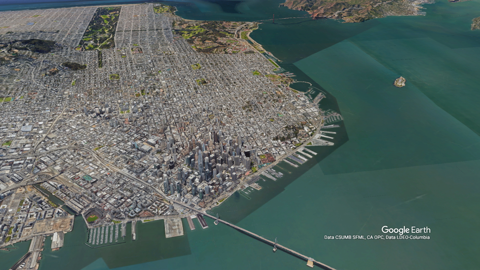}
    \end{subfigure}    
    \begin{subfigure}[b]{0.24\textwidth}    \includegraphics[width=\textwidth]{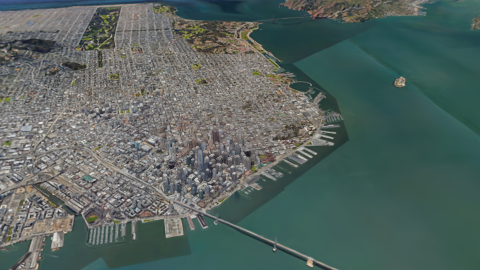}
    \end{subfigure}
\caption{Ground truth vs rendered images of the New York and San Francisco scenes at different altitudes and orientations. Left to right: New York ground truth; New York render; San Francisco ground truth; San Francisco render.} \label{img:NY_SF}
\end{figure*}

As observed in the qualitative comparison in Figure \ref{img:NY_SF}, the images rendered using 3DGS are of high visual quality, and difficult to distinguish visually from ground truth images except for a Google Earth watermark noticeable at the bottom right of ground truth images. Compared with ground truth images, we observe that rendered images have slightly blurrier edges at the smallest scale ($\sim$ 300 m altitude) and that certain street-level details are slightly less sharp at the largest scale ($\sim$ 3000 m altitude). Also noticeable in ground truth Google Earth images and rendered images is the piecing together of multiple data sources at the largest scale at the bottom row of Figure (\ref{img:NY_SF}). We notice visible discontinuous and grid-like change in coloration of the water going from the San Francisco shoreline to the San Francisco Bay and Golden Gate area, likely indicating areas where different aerial and/or satellite images were stitched together. This effect was also learned by the 3DGS model, as is visible in the respective rendered images. 

\begin{table}[H]

\centering
\small
\scalebox{0.85}{\begin{tabular}{l|c|ccc}
\hline
Dataset & Train (PSNR$\uparrow$) & Test (PSNR$\uparrow$ & SSIM$\uparrow$ & LPIPS$\downarrow$) \\ \hline
Waterloo & 32.3 &30.5 & 0.953 & 0.0535\\ \hline
New York & 31.5 & 30.7 & 0.964 & 0.0500 \\
San Francisco & 30.8 & 29.9 & 0.952 & 0.0669 \\ \hline
Amsterdam & 32.3 & 29.7 & 0.948 & 0.0535 \\
Barcelona & 31.2 & 28.1 & 0.937 & 0.0659 \\
Chicago & 32.3 & 30.0 & 0.959 & 0.0460 \\
Los Angeles & 32.0 & 28.6 & 0.914 & 0.0937 \\
Paris & 31.6 & 28.5 & 0.953 & 0.0509 \\
Rome & 32.7 & 27.0 & 0.861 & 0.1127 \\
Quebec & 32.9 & 30.1 & 0.953 & 0.0603 \\
Bilbao & 32.1 & 27.2 & 0.851 & 0.1415 \\ \hline
\end{tabular}}
\caption{Training set PSNR, and Test set PSNR/SSIM/LPIPS results across our scene and various BungeeNeRF scenes.}
\label{tab:psnr}
\end{table}

In addition, we also tested the performance of 3DGS on the other BungeeNeRF Google Earth Studio scenes. We note a 0.7 to 5.7 PSNR drop when moving from training set to test set across all scenes in Table \ref{tab:scenes}, indicating a certain degree of overfitting across training views. We notice the overfitting is more severe on the $\sim 200$ image scenes as opposed to the $\sim 450$ image New York and San Francisco scenes, with our 400 image Waterloo scene lying in the middle. The Bilbao scene, centered on the Gungenheim museum, has by far the worst performing 3DGS reconstructions. We observe that this is perhaps due to a combination of the complex building shape of the Gungenheim museum, a lack of sufficient training views at low altitude, and poorer quality Google Earth 3D model at off-nadir view angles at low altitude which resulted in poor quality training images. 

\begin{table*}[htbp] 
\centering
\begin{tabular}{l|c c c|c c c} 
\hline
 & \multicolumn{3}{c|}{New York (56 Leonard)} & \multicolumn{3}{c|}{San Francisco (Transamerica)} \\ \hline
Method & PSNR$\uparrow$ & LPIPS$\downarrow$ & SSIM$\uparrow$ & PSNR$\uparrow$ & LPIPS$\downarrow$ & SSIM$\uparrow$\\ \hline
NeRF (D=8, Skip=4) \citep{2021nerf} & 21.7 & 0.320 & 0.636 & 22.6 & 0.318 & 0.690 \\
NeRF w/ WPE (D=8, Skip=4) \citep{2021nerf}& 21.6 & 0.365 & 0.633 & 22.4 & 0.331 & 0.680 \\
Mip-NeRF-small (D=8, Skip=4) \citep{2021mipnerf} & 22.0 & 0.344 & 0.648 & 22.7 & 0.327 & 0.687 \\
Mip-NeRF-large (D=10, Skip=4) \citep{2021mipnerf}& 22.2 & 0.318 & 0.666 & 22.5 & 0.330 & 0.686 \\
Mip-NeRF-full (D=10, Skip=4,6,8) \citep{2021mipnerf}& 22.3 & 0.266 & 0.689 & 22.8 & 0.314 & 0.699 \\
BungeeNeRF (same iter as baselines) \citep{2022bungeenerf}& 23.5 & 0.235 & 0.739 & 23.6 & 0.265 & 0.749 \\
BungeeNeRF (until convergence) \citep{2022bungeenerf}& 24.5 & 0.160 & 0.815 & 24.4 & 0.192 & 0.801 \\ \hline
3DGS & \textbf{30.7} & \textbf{0.050}  & \textbf{0.964} & \textbf{29.9} & \textbf{0.067} & \textbf{0.952} \\ \hline
\end{tabular}
\caption{PSNR, LPIPS, and SSIM benchmark on BungeeNeRF New York and San Francisco scenes, NeRF results cited from \citep{2022bungeenerf}.}
\label{tab:2sceneBenchmark}
\end{table*}

\subsection{3D Reconstruction of the Region of Study}
\begin{figure*}[h!]
    \centering
    \begin{subfigure}[b]{0.32\textwidth}
\includegraphics[width = 1\textwidth, trim={10cm 7cm 10cm 3cm},clip]{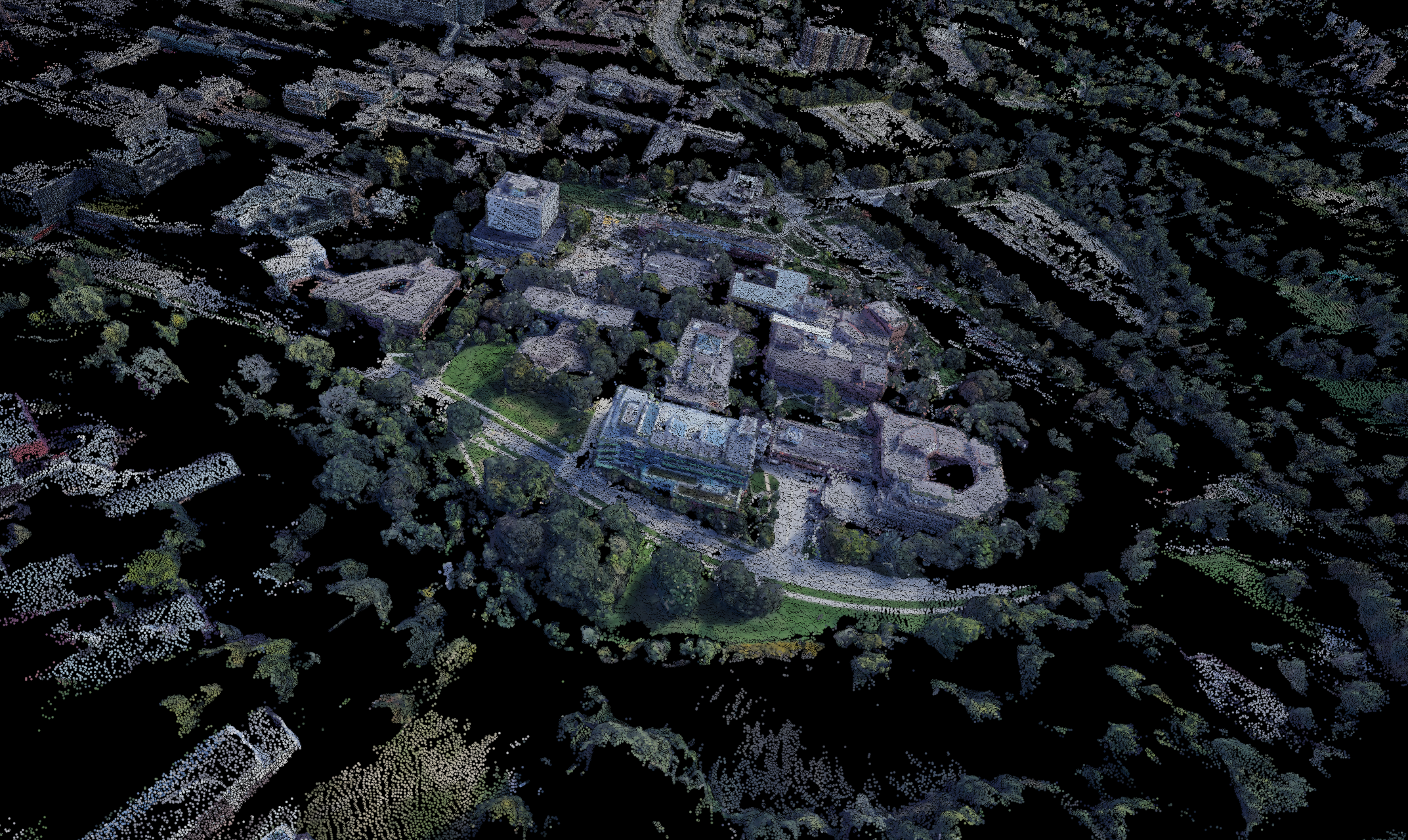}
    \end{subfigure}   
    \begin{subfigure}[b]{0.32\textwidth}
\includegraphics[width = 1\textwidth, trim={10cm 7cm 10cm 3cm},clip]{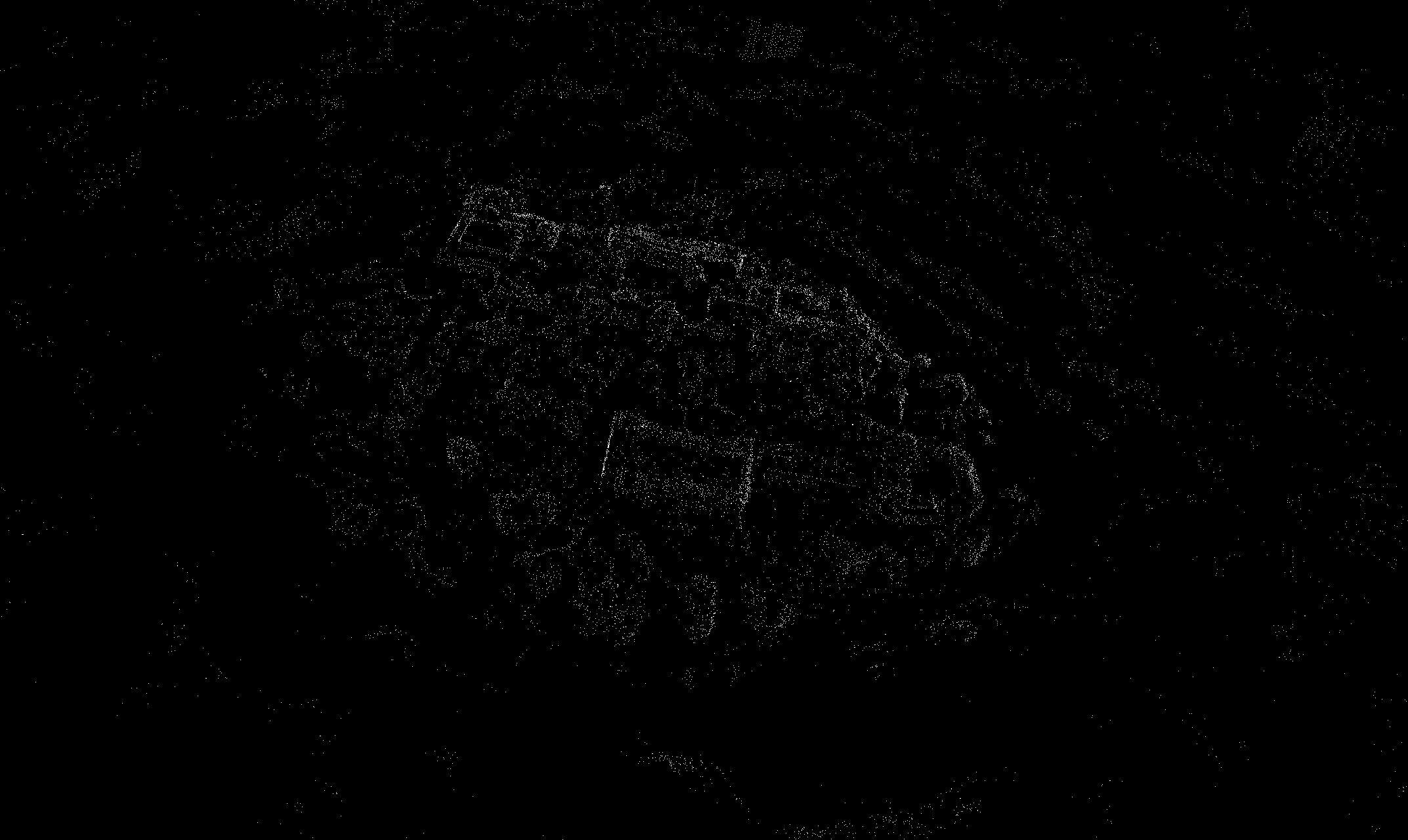}
    \end{subfigure}~
    \begin{subfigure}[b]{0.32\textwidth}
\includegraphics[width = 1\textwidth, trim={10cm 7cm 10cm 3cm},clip]{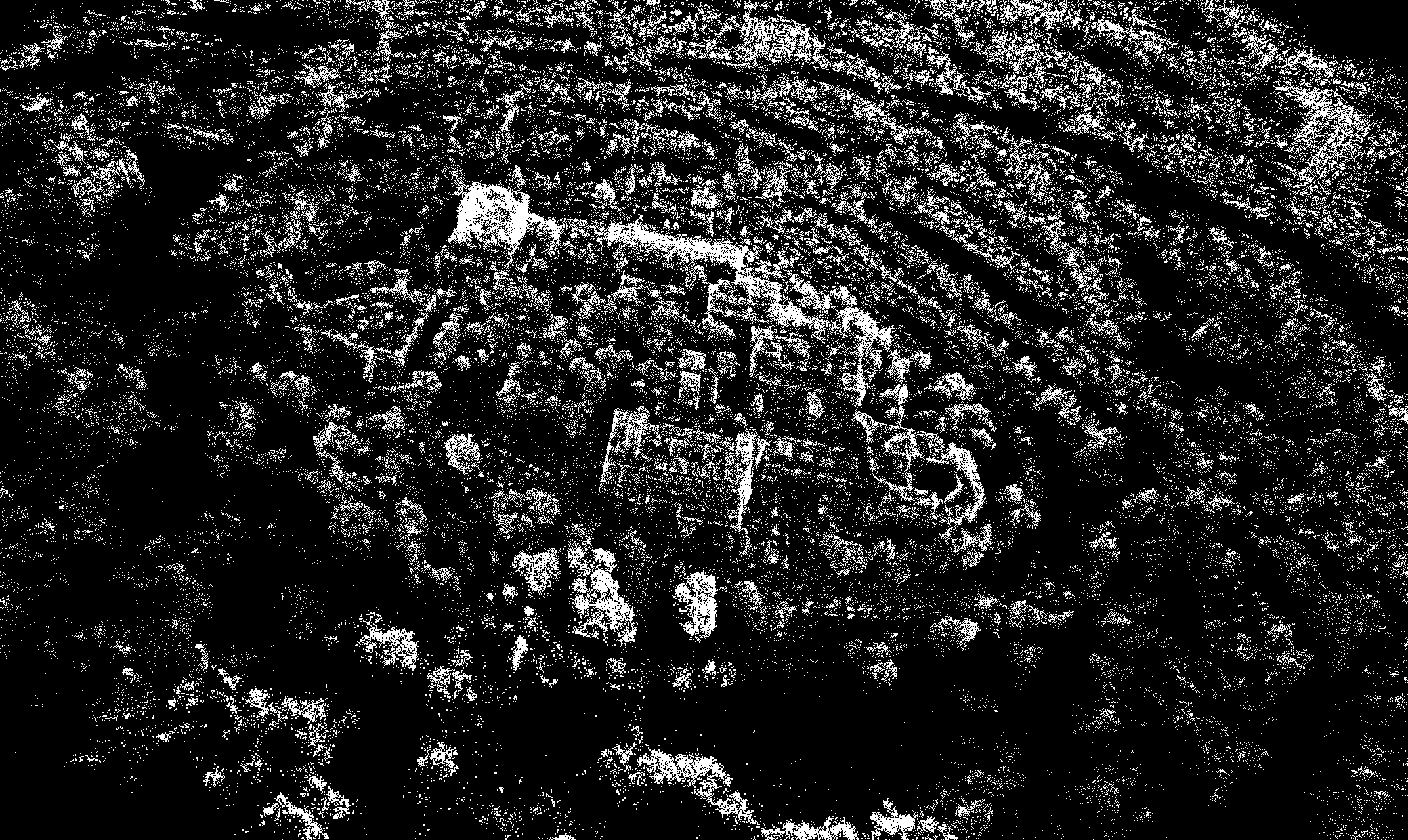}
    \end{subfigure}    
    
    \centering
    \begin{subfigure}[b]{0.32\textwidth}
\includegraphics[width = 1\textwidth, trim={10cm 7cm 10cm 3cm},clip]{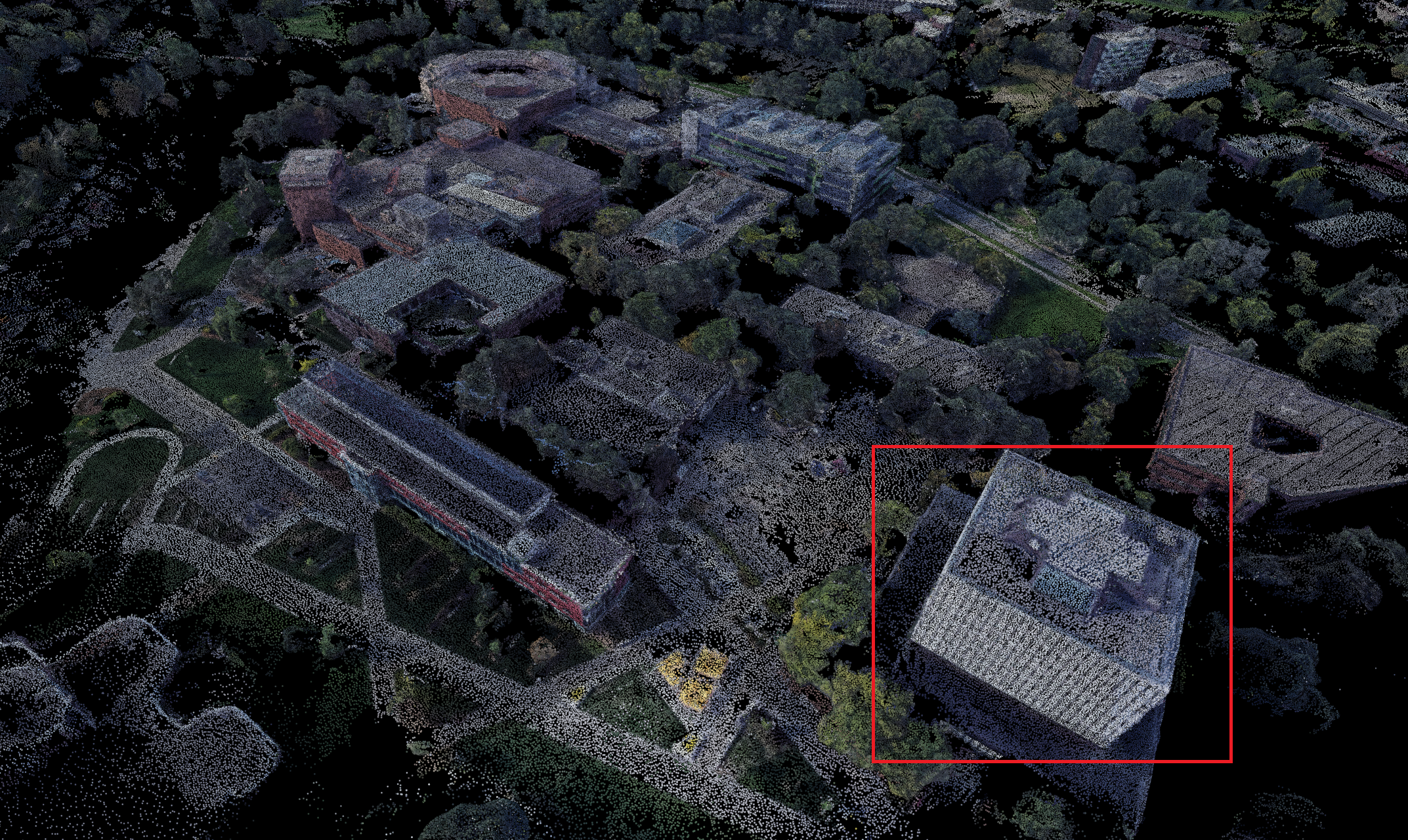}
    \end{subfigure}   
    \begin{subfigure}[b]{0.32\textwidth}
\includegraphics[width = 1\textwidth, trim={10cm 7cm 10cm 3cm},clip]{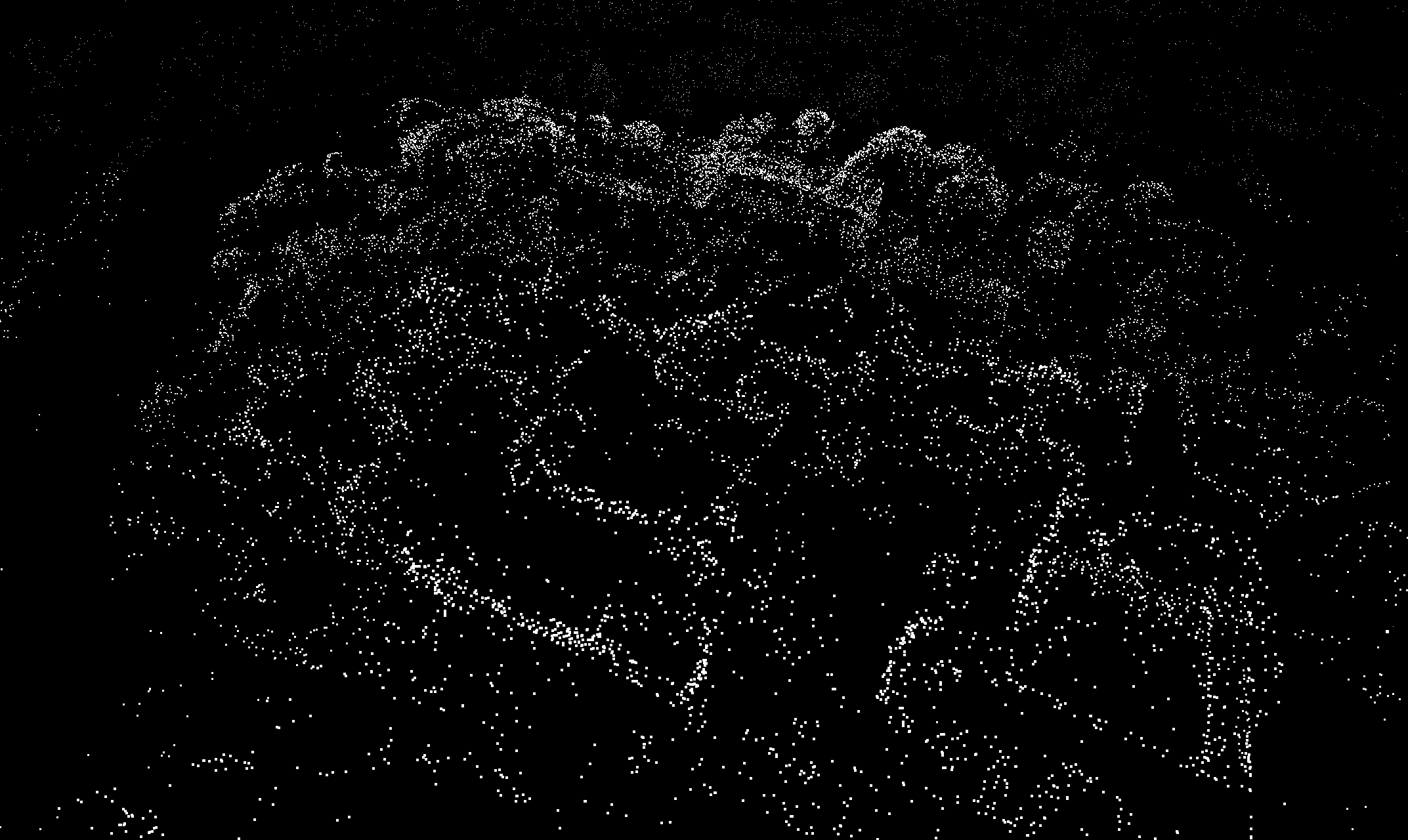}
    \end{subfigure}~
    \begin{subfigure}[b]{0.32\textwidth}
\includegraphics[width = 1\textwidth, trim={10cm 7cm 10cm 3cm},clip]{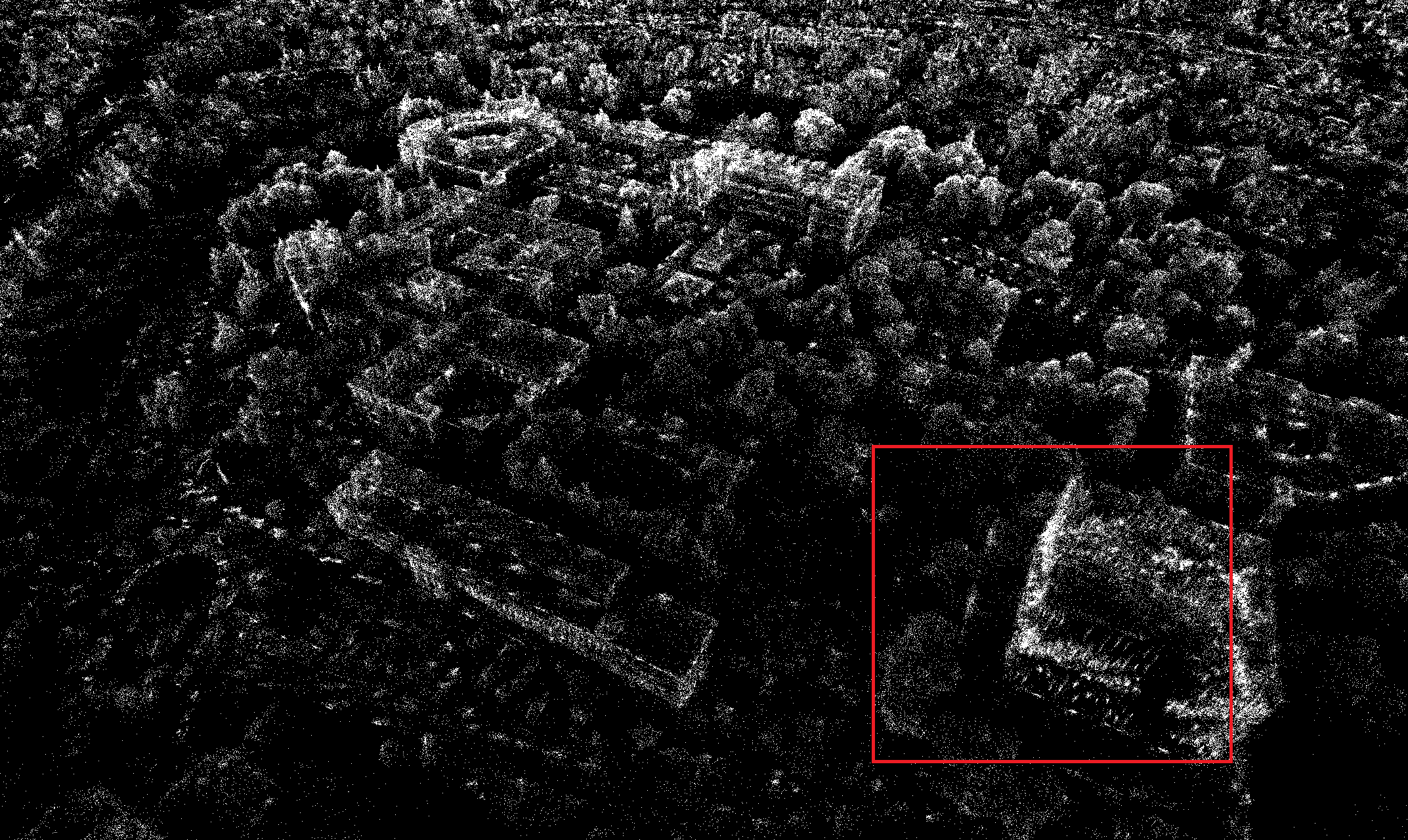}
    \end{subfigure}    

    \centering
\begin{subfigure}[b]{0.32\textwidth}
\includegraphics[width = 1\textwidth, trim={10cm 8cm 10cm 2cm},clip]{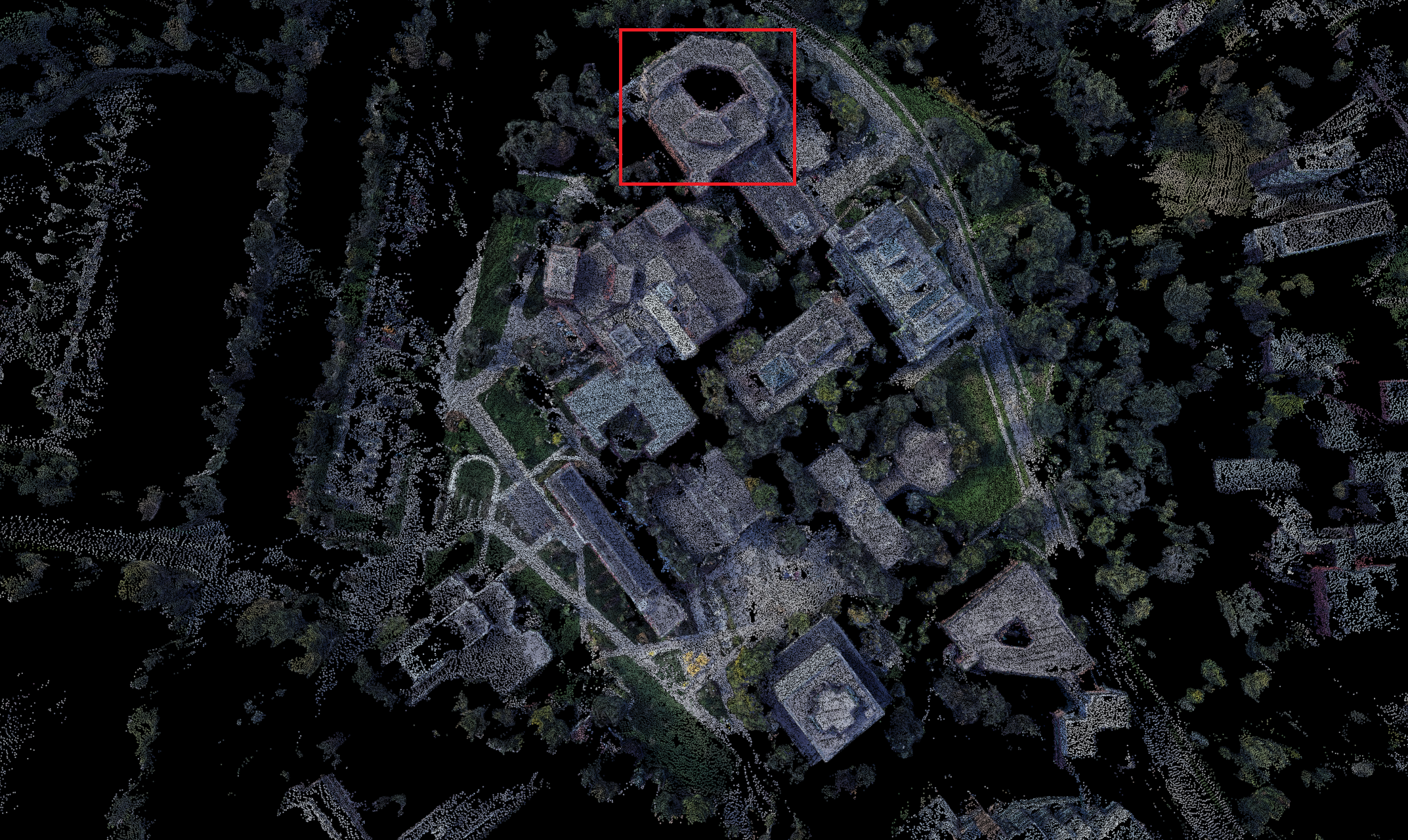}
    \end{subfigure}   
    \begin{subfigure}[b]{0.32\textwidth}
\includegraphics[width = 1\textwidth, trim={10cm 8cm 10cm 2cm},clip]{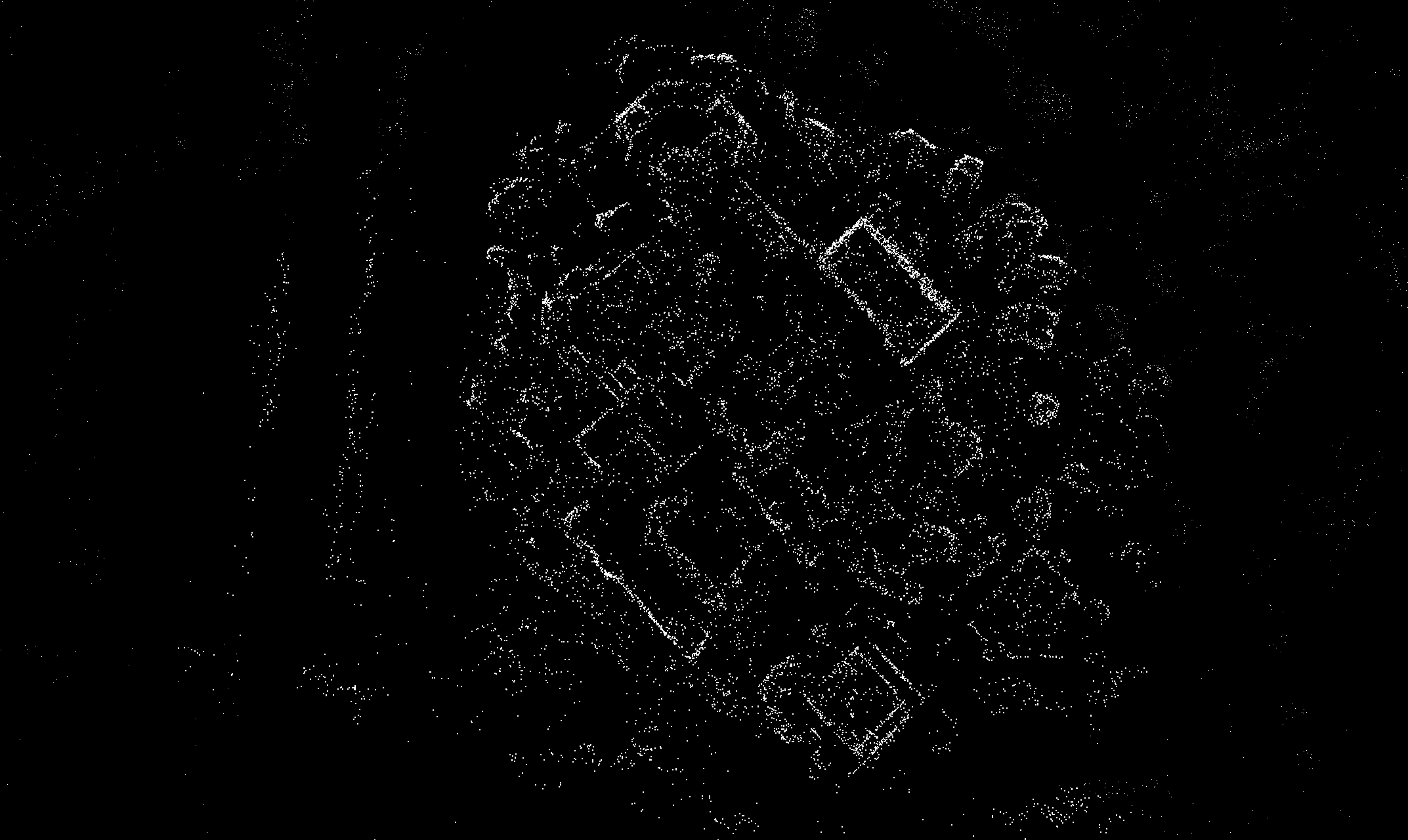}
    \end{subfigure}~
    \begin{subfigure}[b]{0.32\textwidth}
\includegraphics[width = 1\textwidth, trim={10cm 8cm 10cm 2cm},clip]{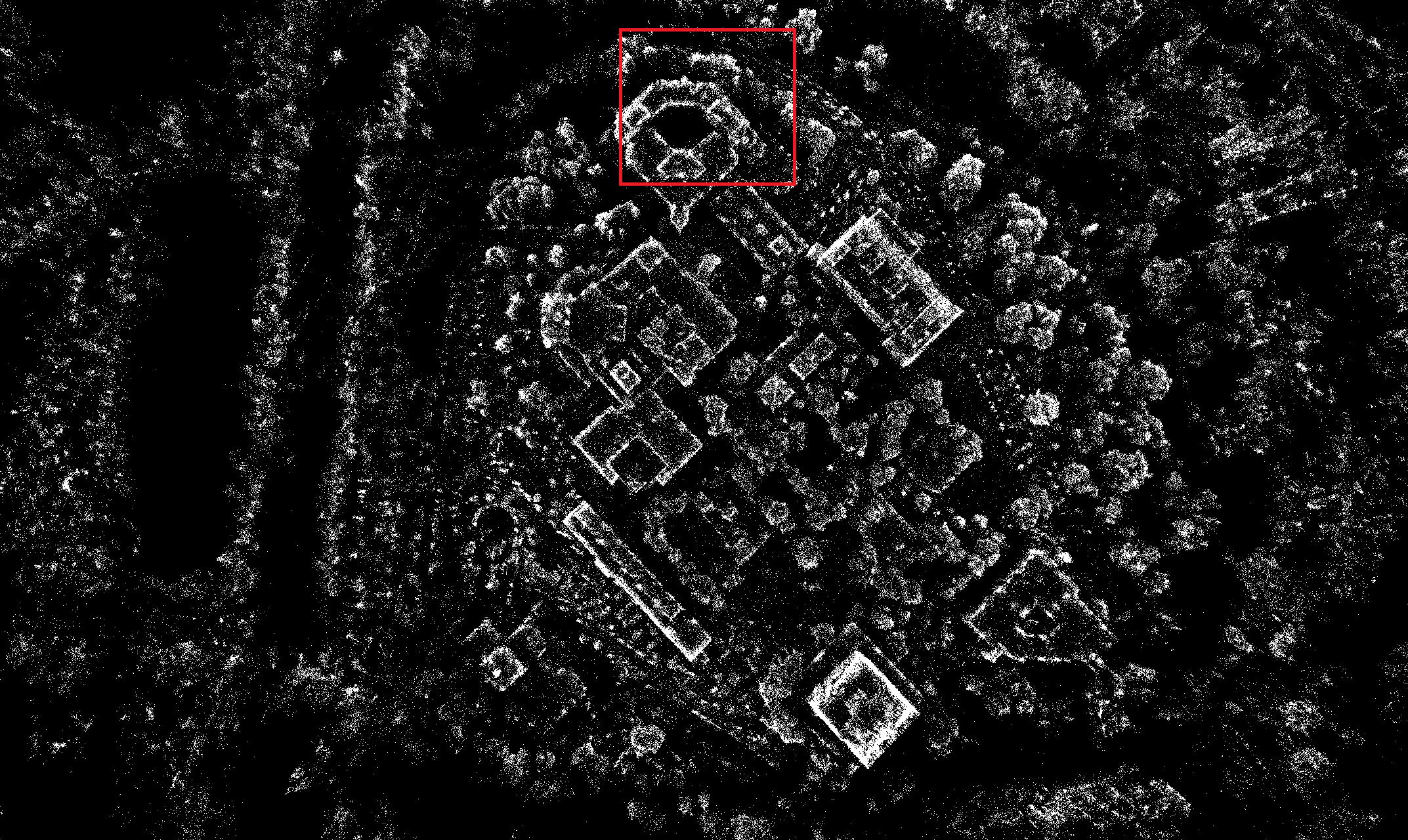}
    \end{subfigure}

\caption{Comparison of 3DGS point cloud densification of the EV-1 neighborhood at the University of Waterloo against initial and MVS densified point clouds. Left: Multi-view stereo depth and normal maps fusion as ground truth (colored) dense point cloud. Middle: initial sparse point cloud. Right: 3DGS densified point cloud. Areas with obvious discrepancies/misalignment are highlighted in red rectangles (aligned-pixelwise). } \label{img:pc_compare}
\end{figure*}

For the 3D reconstruction experiments, due to computational constraints of multi-view stereo (MVS) densification, which is even more memory intensive than 3DGS, we extracted the sparse point cloud using the first 50 images along the first level camera path. Then, using \cite{2016mvs}, we generated depth and normal maps, which we then used to generate a dense point cloud as the ground truth/reference 3D geometry of the EV1 neighborhood. We then trained a 3D Gaussian Splatting model on these first 50 images. After which, the Gaussian positional means were extracted as a new 3DGS densified 3D point cloud resulting in 1856968 points, starting from a sparse point cloud of 24740 points, a near 10x densification. The positional means which we extracted as 3DGS densified point cloud are visualized in Figure \ref{img:uw}(rasterized at Gaussian scale = $10^{-3}$). In comparison, the MVS densified point cloud resulted in 2528969 points. The MVs densification results are visualized in Figure \ref{img:pc_compare}. 

\begin{figure*}[H]
    \centering
    \begin{subfigure}[b]{0.32\textwidth}
\includegraphics[width = 1\textwidth]{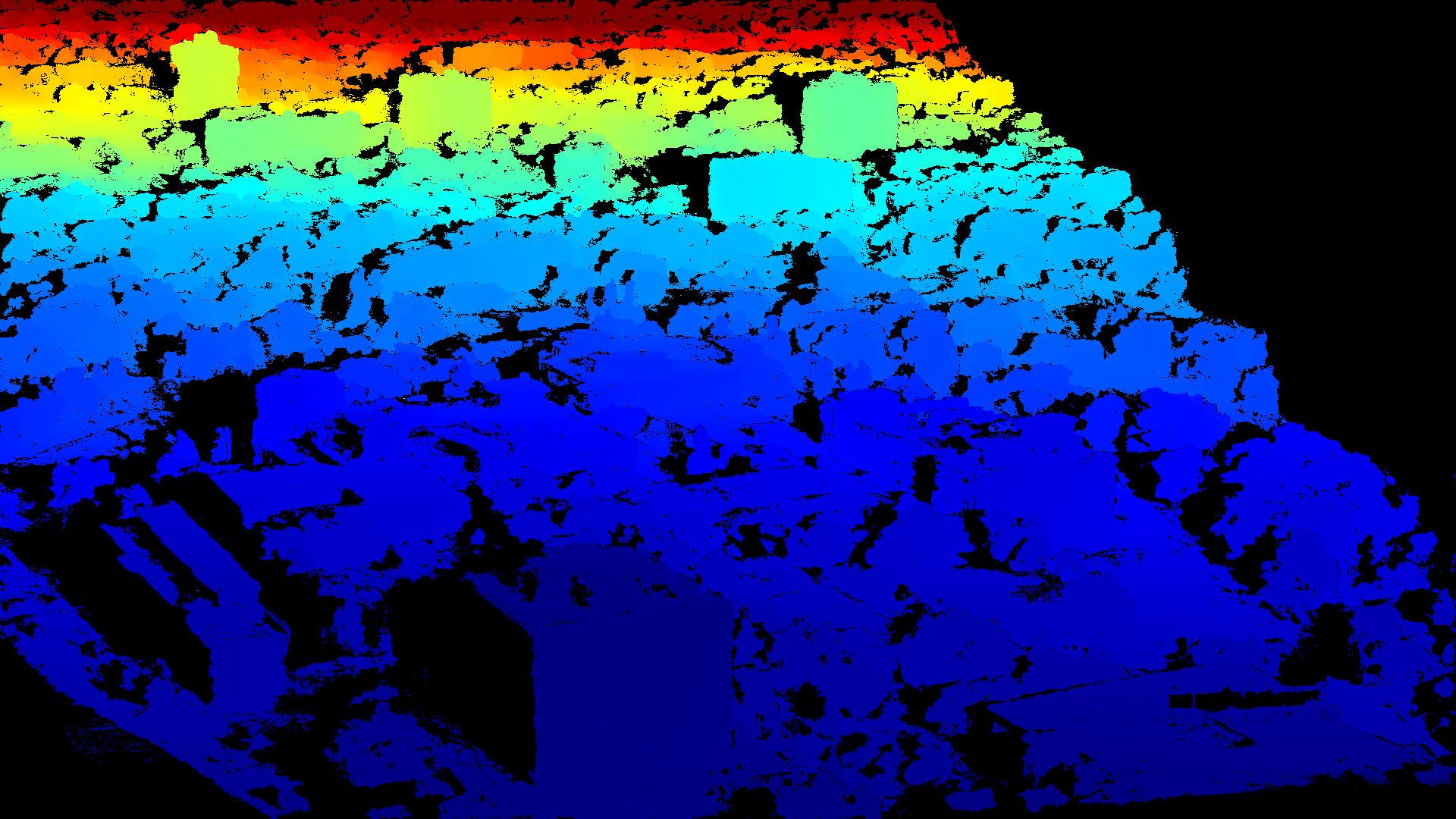}
    \end{subfigure}       
    \begin{subfigure}[b]{0.32\textwidth}
\includegraphics[width = 1\textwidth]{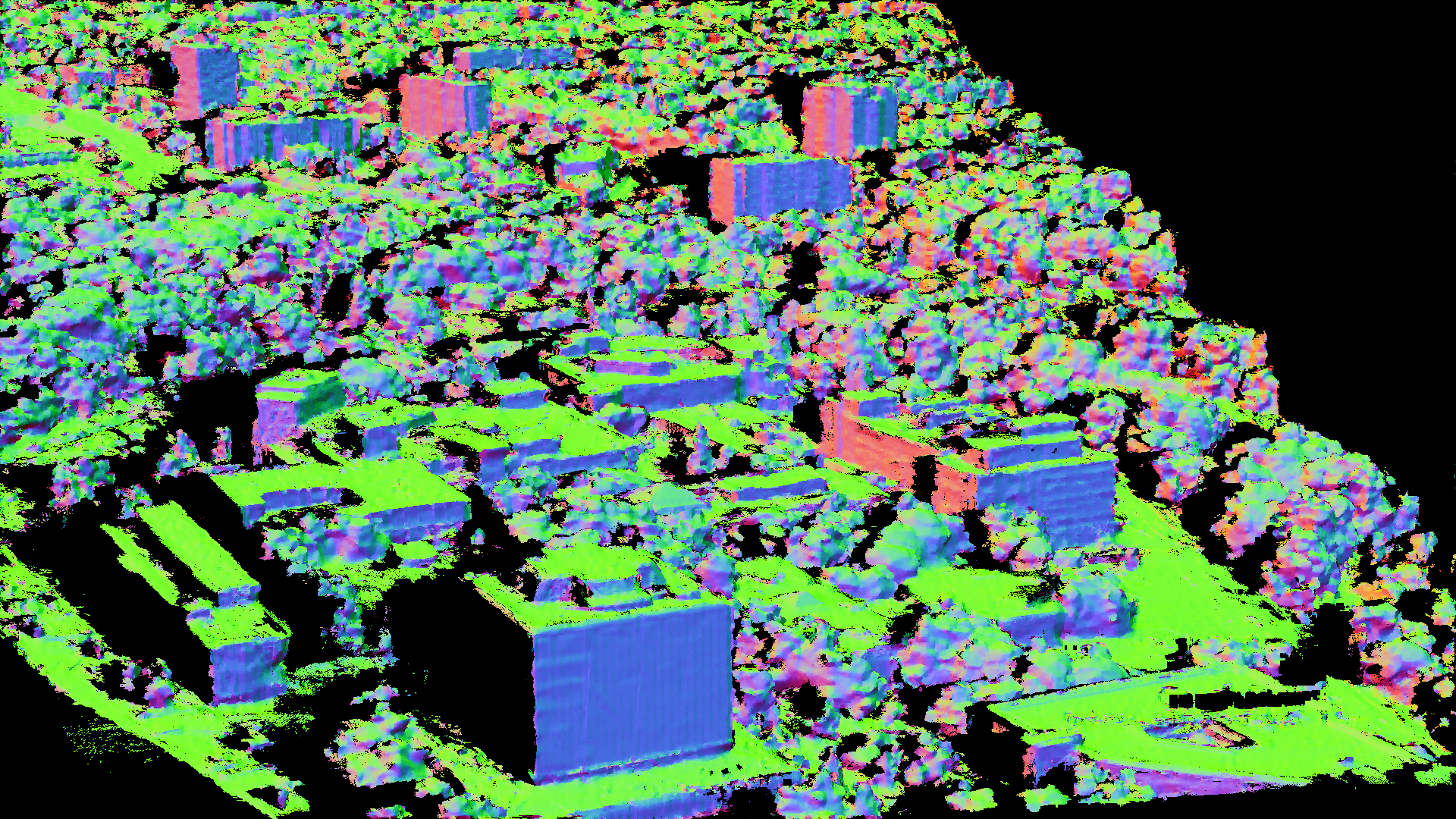}
    \end{subfigure}    
    \begin{subfigure}[b]{0.32\textwidth}
\includegraphics[width = 1\textwidth]{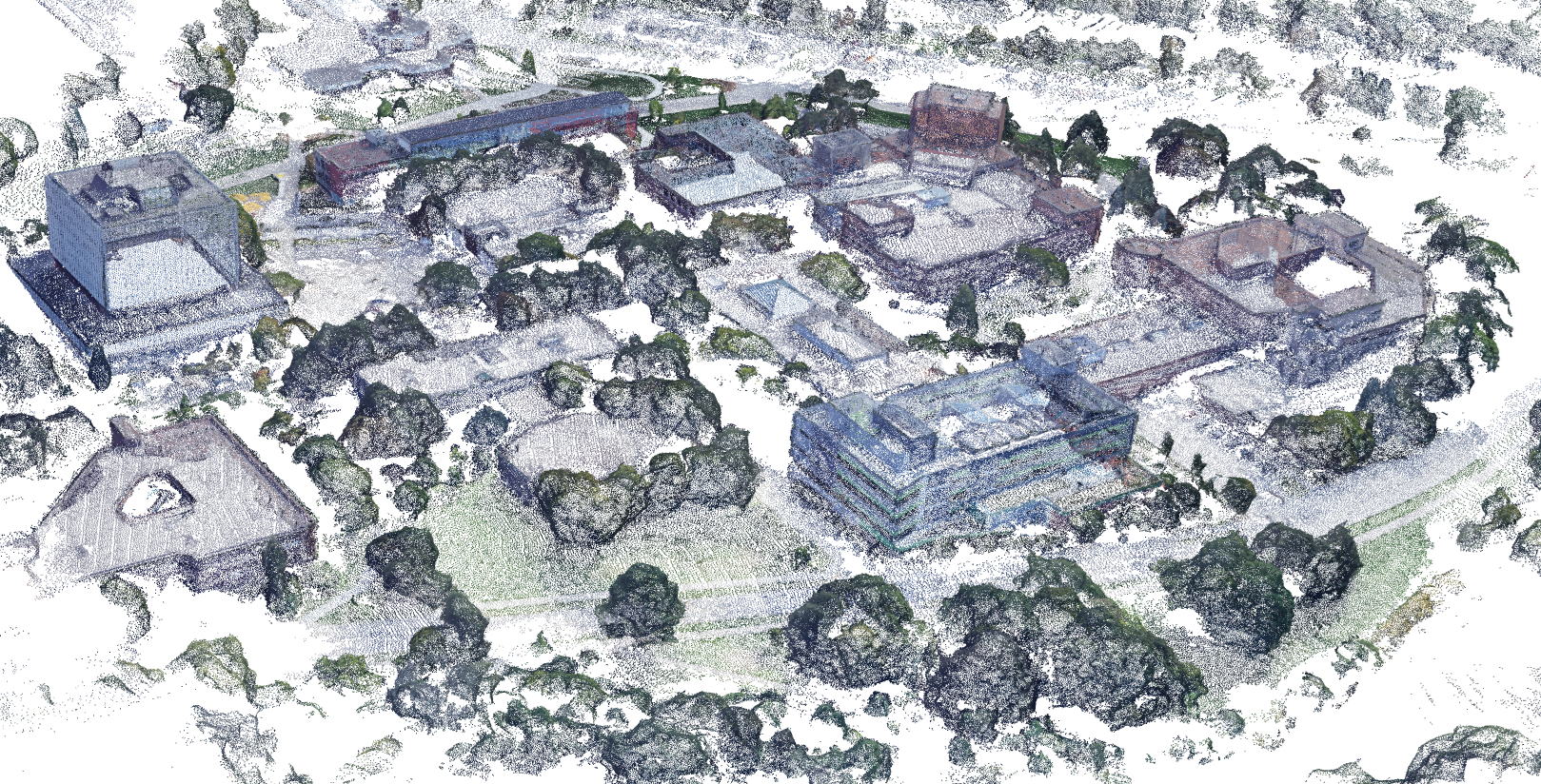}
    \end{subfigure}
    \caption{Multi-View-Stereo densification. Left: Example depth map. Middle: Example normal map. Right: MVS reconstructed dense colored point cloud.} \label{img:normal,depth}
\end{figure*}

\begin{figure*}[H]
    \centering
    \begin{subfigure}[b]{0.32\textwidth}
\includegraphics[width = 1\textwidth, trim={8cm 4cm 9cm 2cm},clip]{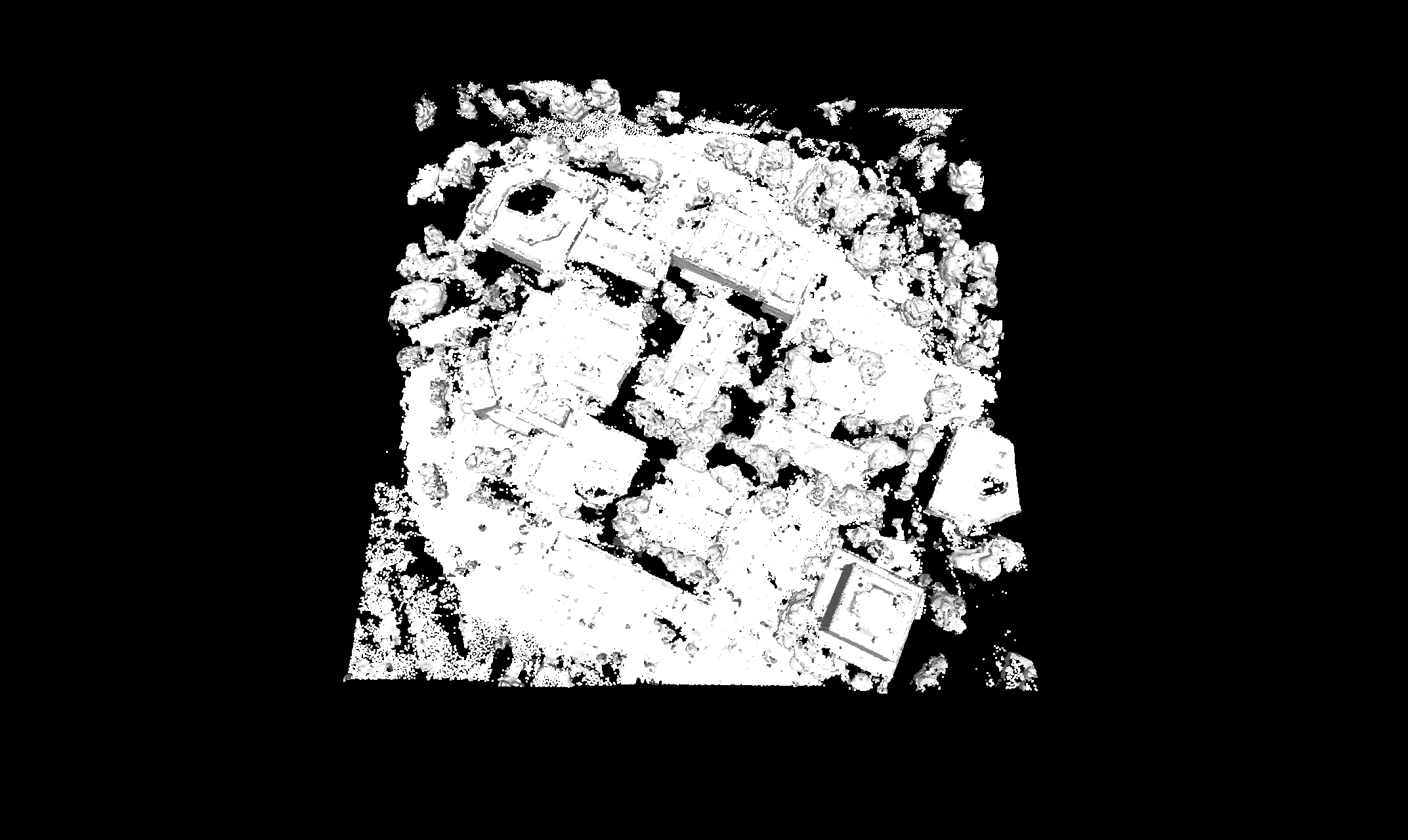}
    \end{subfigure}   
    \begin{subfigure}[b]{0.32\textwidth}
\includegraphics[width = 1\textwidth, trim={8cm 4cm 9cm 2cm},clip]{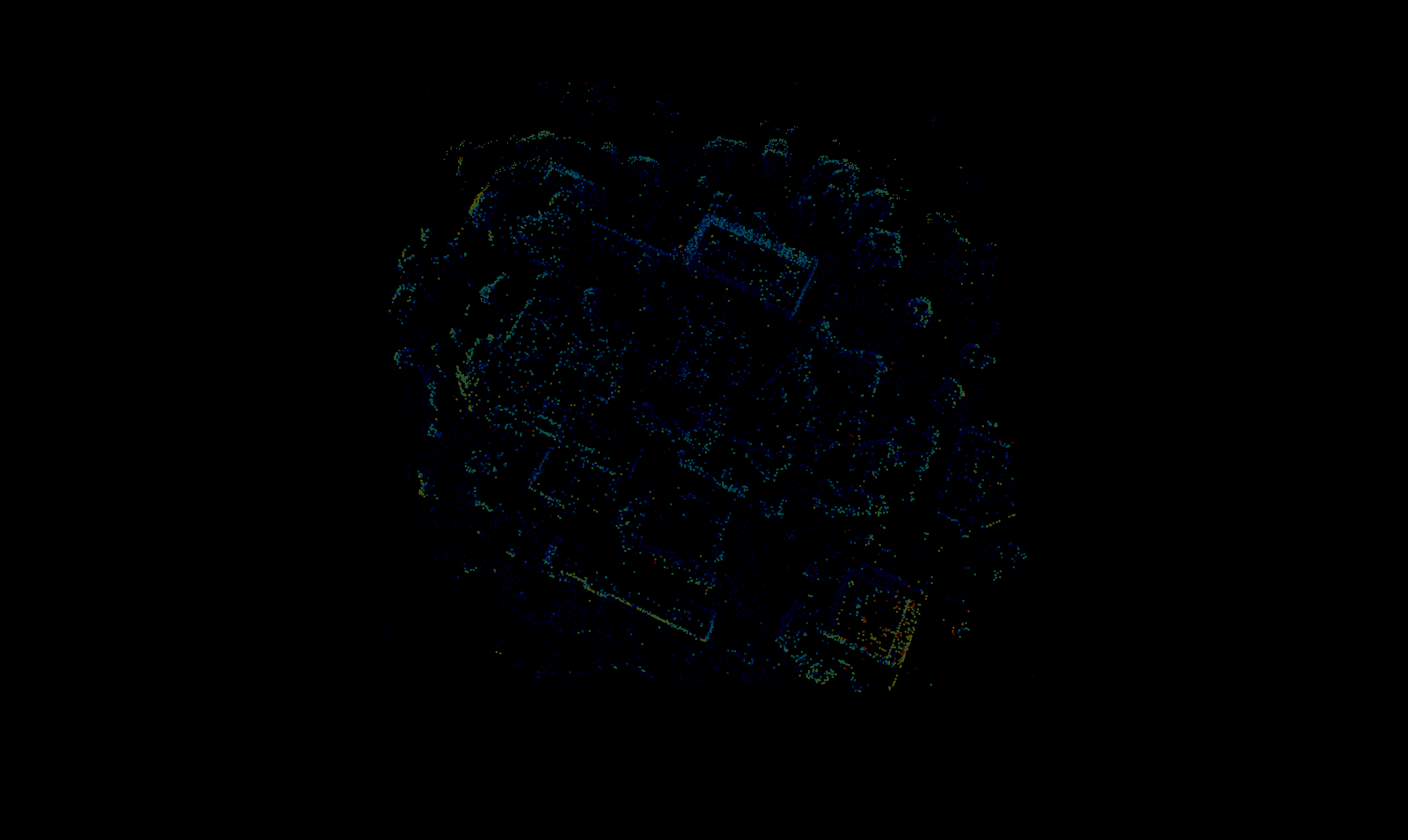}
    \end{subfigure}
    \begin{subfigure}[b]{0.32\textwidth}
\includegraphics[width = 1\textwidth, trim={8cm 4cm 9cm 2cm},clip]{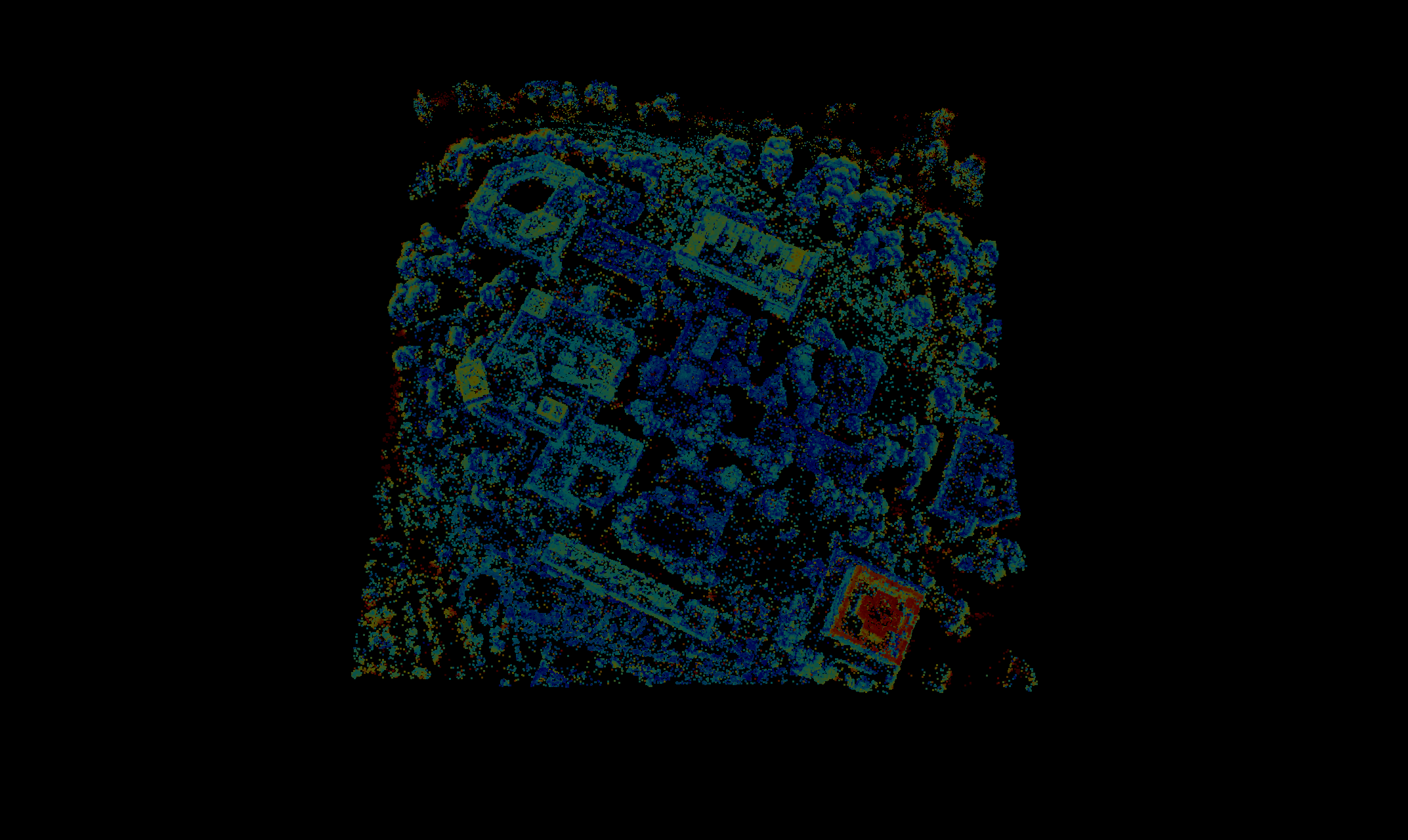}
    \end{subfigure}

    \begin{subfigure}[b]{0.32\textwidth}
\includegraphics[width = 1\textwidth, trim={8cm 4cm 9cm 2cm},clip]{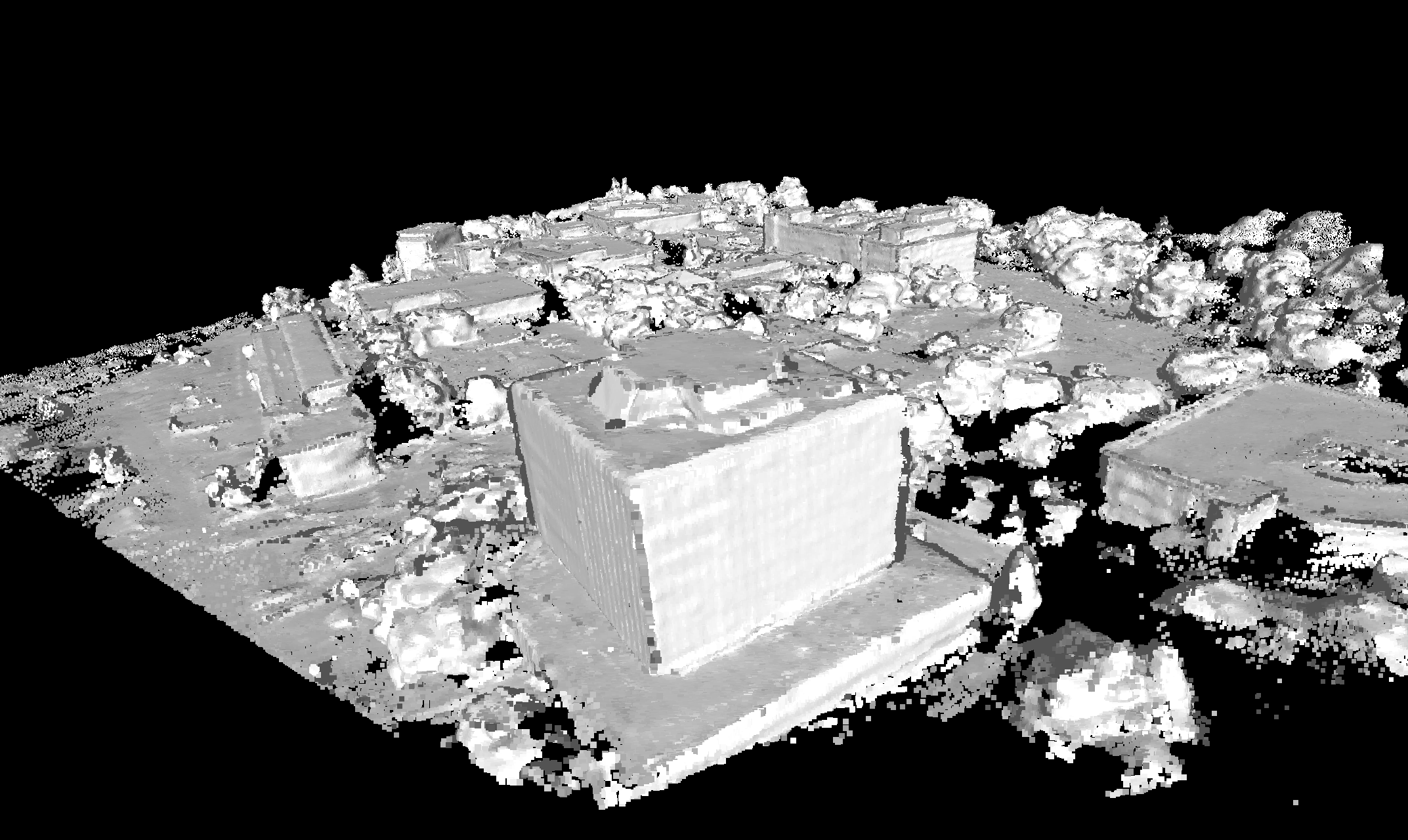}
    \end{subfigure}   
    \begin{subfigure}[b]{0.32\textwidth}
\includegraphics[width = 1\textwidth, trim={8cm 4cm 9cm 2cm},clip]{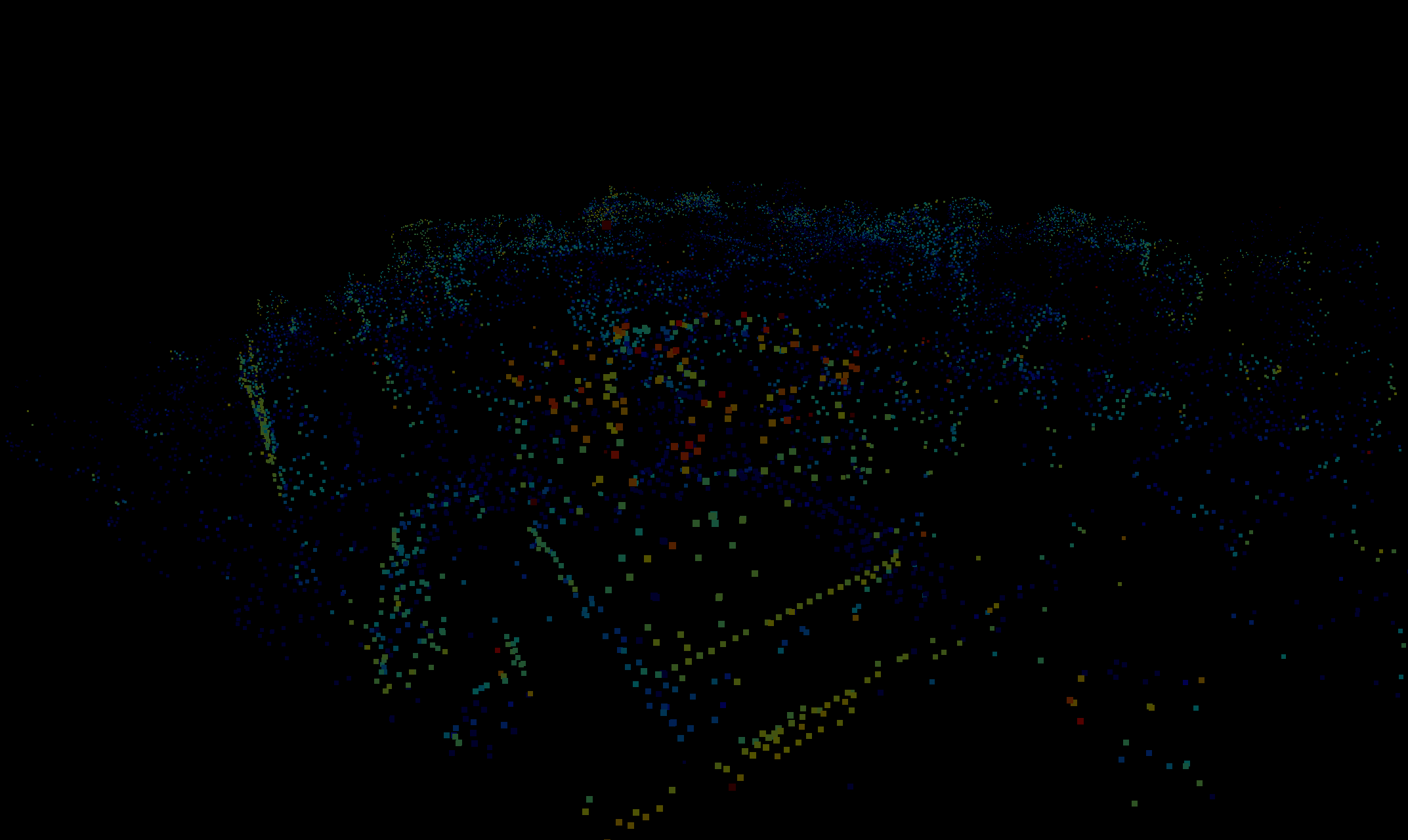}
    \end{subfigure}
    \begin{subfigure}[b]{0.32\textwidth}
\includegraphics[width = 1\textwidth, trim={8cm 4cm 9cm 2cm},clip]{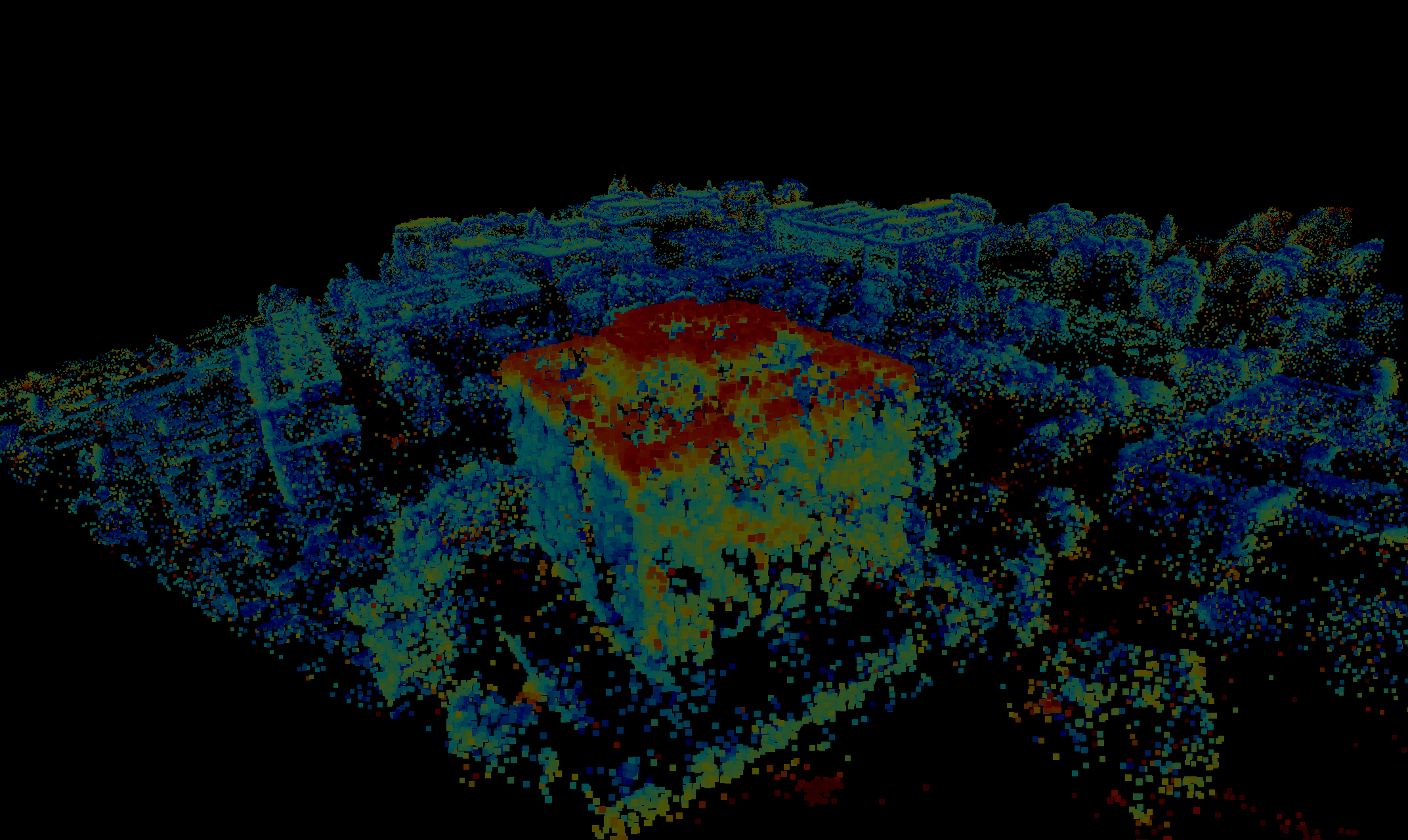}
    \end{subfigure}

    \caption{Post cropping and registration visualization of reference point cloud and deviation in local Hausdorff distance from reference (blue = small deviation, red = large deviation). Left: reference MVS densified point cloud, meshed using Poisson surface reconstruction \citep{2013poisson}. Middle: sparse initial point cloud colored by Hausdorff distance from reference mesh.
    Right: 3DGS densified point cloud colored by Hausdorff distance from reference mesh.} \label{img:crop}
\end{figure*}

We first note that the initial sparse point cloud and the 3DGS densified point clouds are aligned with each other. However, the MVS densified point cloud which we take to be ground truth/reference point cloud was offset from both by a rotation, a translation, and non-affine deformations far from the origin. This is visible in the last row of Figure \ref{img:pc_compare}, but even more obvious as we extend the view further. 
As such, we cropped all three point clouds, and performed point cloud registration to align the three point clouds. We used the iterative closest point algorithm (ICP) \citep{1992icp} to register both the initial and the 3DGS densified point clouds to the MVS densified point cloud. This process aligned all three point clouds with translations and rotations, but we still observe slight non-affine deformations as distance from origin and height increases post registration, as can be seen in Figure \ref{img:crop}. 

The cropping resulted in 12773, 244849, 1270820, points for the sparse, 3DGS densified, and MVS densified point clouds respectively. We noticed that both the initial sparse point cloud and the MVS densified point cloud were much denser at the center of the scene than at the edges, whereas the 3DGS densified point cloud had proportionately a more uniform point density compared to to the two aforementioned point clouds. As such, the cropping reduced the number of points in the sparse and MVS densified point clouds by roughly a factor of $\sim$2, whereas it reduced the number of points in the 3DGS densified point cloud by a factor of $\sim$7.5. 

 We then compared both point clouds to the dense MVS point cloud fused from depth and normal maps, using D1 (point-to-point) MSE and D2 (point-to-surface) MSE, Haussdorff distance, and Chamfer distance. We observe that the 3DGS densified point cloud has marginally higher D1 and D2 MSE (with respect to the MVS densified point cloud) than the sparse initial point cloud. However, we note that neither MSE metric penalizes differences in point density. They only measure the presence of outliers and noise points. On the other Haussdorff and Chamfer distances better reflect the differences between distribution of points. We observe that compared to the sparse point cloud, the 3DGS densified point cloud has much better better agreement with respect to the reference MVS densified point cloud in terms of these two metrics. This is also corroborated with visual inspection of Figure \ref{img:pc_compare}. We plotted the local Hausdorff distance with respect to the reference MVS densified point cloud in Figure \ref{img:crop}. Which helped highlight the non-affine distortion between the reference MVS densified point cloud and the two others.

\begin{table}[htbp]
\centering
\caption{Comparison of initial sparse point cloud and 3DGS densified point cloud vs. MVS densified point cloud as reference.}
\label{tab:sparse_vs_3dgs}
\begin{tabular}{@{}l|rr@{}}
\hline
 & Sparse & 3DGS Densified \\ \hline
Points & 24740 & 1856968 \\
Points post-cropping & 12773 & 244849 \\
\hline
D1 MSE$\downarrow$ & $7.625 \times 10^{-3}$ & $8.154\times 10^{-3}$ \\
D2 MSE$\downarrow$ & $6.879\times 10^{-3}$ & $7.297 \times 10^{-3}$ \\
Haussdorff distance$\downarrow$ & $8.753\times 10^{-1}$ & $3.745\times 10^{-1}$ \\
Chamfer distance$\downarrow$ & $2.546\times 10^{-2}$ & $1.615\times10^{-2}$ \\ \hline
\end{tabular}
\end{table}


\section{Discussions}

We note that Google Earth Studio produces composite images and images rendered from 3D models, constructed using remote sensing images of a variety of governmental and commercial sources including Landsat, Copernicus, Airbus, NOAA, U.S. Navy, USGS, Maxar, taken at different times. This can be both an advantage and a disadvantage. We first note that low-altitude images using far-from-vertical/off-nadir point-of-views rely on Google Earth Engine's own 3D models, which are limited in detail compared to real remote sensing images. On the other hand, the variety of data sources benefits the robustness of the 3D Gaussian Splatting model, which has been trained on images from different sensors in different photometric and radiometric conditions. The disadvantages are also counterbalanced with the ease at which Google Earth Studio allows for the creation of multi-scale dataset with spiraling camera path suited for large-scale 3D scene centered around some neighborhood of interest in a city.

When recovering 3D geometry (as 3D point cloud) from both the SfM preprocessing, the post-3DGS densified point cloud, and even the MVS densified dense point cloud, we notice a mild to strong presence of noise, which should be address in future 3DGS research. In our 3D reconstruction and densification experiments, we used the MVS densified point cloud as ground truth, despite that it was also constructed from 2D images. Despite recovering good quality dense 3D surfaces, the MVS densified point cloud was offset from both the initial sparse point cloud and the 3DGS densified point cloud with a non-affine transformation, which should be investigated further. In the future, for geometry recovery benchmarks, we believe a scanned point cloud such as from a LiDAR source would be more accurate as ground truth. A future project could be to use a scanned point cloud as ground truth and to register and georeference both MVS and 3DGS densified point clouds to a scanned point cloud to properly study the geometry of these densifications, as well as enable further mapping and GIS applications. We also note that memory requirement for the COLMAP MVS densification was larger than that of the 3DGS densification, and was a reason behind performing the densification experiment on a smaller scale using less images. Despite these concerns, we note despite that 3DGS was not built as a 3D geometry extraction tool, it is reasonably able to recover scene geometry through densification and optimization of the Gaussian positions. 

The high GPU memory requirement of 3D Gaussian Splatting prevents high resolution reconstruction across the entire large-scale scene. Due to the chosen camera path, the center of the scene is well reconstructed at all altitudes and densely populated by Gaussians. This results in high quality rendered images. However, in other neighborhoods further away from the scene center, we are only able to achieve high quality reconstruction at high altitude and struggle near the ground. Although certain advances, currently in preprints, have attempted to address the memory issue; many of these models compress the trained Gaussian Splatting model post-training reducing model storage requirements, and do not achieve significant reduction in working memory requirement during training.

We expect working memory requirement reduction to be a future research direction. This will also allow for better reconstruction across multiple neighborhoods, perhaps using more complex camera paths such as multiple spirals arranged hierarchically in Google Earth Studio, centered around each neighborhood of interests, or space-filling curves filling out a camera path with dense coverage across the entire large-scale scene. Alternatively, large-scale 3D reconstruction scheme which pieces together multiple local models such as with Mega-NeRF \citep{2022mega} can be considered. Another future research direction is remote sensing-based large-scale semantics-based 3D reconstruction and semantic synthesis. For urban scenes, this research area is expected to find applications in urban digital twin creation, urban monitoring, and urban/land-use planning. This research direction can also more generally extend land-use/land-cover segmentation to three dimensions, which has a multitude of research and commercial applications. These are the research areas we are currently investigating.
\section{Conclusion}
\label{sec:conclusion}

By simply leveraging Google Earth imagery we capture an aerial off-nadir dataset of the region of study. We were able to photorealistically render the scene and capture its geometry. We compared the 3DGS with NeRF methods on a large-scale urban reconstruction dataset across 10 cities, and performed a careful study of the 3D point cloud densification capability of 3DGS comparing and visualizing the densification against Multi-View-Stereo dense reconstruction in our region of study. We find both an affine misalignment which we remove with a point cloud registration and a non-linear deformation which we quantify and visualize between the Multi-View-Stereo densified point cloud and 3DGS densified point cloud. We hope our study and experiments help future research in large-scale remote sensing-based 3D Gaussian Splatting for both view synthesis and geometry retrieval.

\printcredits

\section*{Declaration of competing interest}
The authors declare that they have no known competing financial interests or personal relationships that could have appeared to influence the work reported in this paper.

\section*{Acknowledgements}

\bibliographystyle{cas-model2-names}
\bibliography{11_references}

\end{document}